\documentclass[times, review, 10pt]{elsarticle}
\usepackage{setspace}
\usepackage{amssymb}
\usepackage{amsmath}
\usepackage{algorithm}
\usepackage{algpseudocode} 
\usepackage{amsmath} 
\usepackage{float} 
\usepackage{graphicx} 
\usepackage{subcaption} 
\usepackage{booktabs} 
\usepackage{multirow} 
\usepackage{hyperref} 
\usepackage{enumitem}
\usepackage{microtype} 
\usepackage[export]{adjustbox}
\usepackage{color}
\journal{Pattern Recognition}

\begin{document}

\begin{frontmatter}

\title{Hybrid second-order gradient histogram based global low-rank sparse regression for robust face recognition}

\author[label1]{Hongxia Li}
\ead{pippx81@163.com}

\author[label2,label1]{Ying Ji}
\ead{425358837@qq.com}

\author[label3]{Yongxin Dong}
\ead{dy893291318@163.com}

\author[label1]{Yuehua Feng\corref{correspondingauthor}}
\cortext[correspondingauthor]{Corresponding author}
\ead{yhfeng@sues.edu.cn}

\address[label1]{School of Mathematics, Physics and Statistics, Shanghai University of Engineering Science, Shanghai, 201620, China.}
\address[label2]{School of Mathematical Sciences , Sunway University, Kuala Lumper 47500, Malaysia.}
\address[label3]{School of Law and Criminal Justice, East China University of Political Science and Law, Shanghai, 201620, China.}


\begin{abstract}
Low-rank sparse regression models have been widely adopted in face recognition due to their robustness against occlusion and illumination variations. However, existing methods often suffer from insufficient feature representation and limited modeling of structured corruption across samples. To address these issues, 
this paper proposes a Hybrid second-order gradient Histogram based Global Low-Rank Sparse Regression (H2H-GLRSR) model. First, we propose the Histogram of Oriented Hessian (HOH) to capture second-order geometric characteristics such as curvature and ridge patterns. 
By fusing HOH and first-order gradient histograms, we construct a unified local descriptor, termed the Hybrid second-order gradient Histogram (H2H), which enhances structural discriminability under challenging conditions. Subsequently, the H2H features are incorporated into an extended version of the Sparse Regularized Nuclear Norm based Matrix Regression (SR\_NMR) model, where a global low-rank constraint is imposed on the residual matrix to exploit cross-sample correlations in structured noise. The resulting H2H-GLRSR model achieves superior discrimination and robustness. Experimental results on benchmark datasets demonstrate that the proposed method significantly outperforms state-of-the-art regression-based classifiers in both recognition accuracy and computational efficiency.
\end{abstract}








\end{frontmatter}

\section{Introduction}
Face recognition continues to be one of the most important and dynamic research areas in computer vision, with applications ranging from biometrics and security surveillance to human-computer interaction and social media \cite{zhao2003face, jain2011handbook, wang2021deep}. Over the past decades, a multitude of face recognition methods have achieved significant progress; however, challenges remain when dealing with complex variations in images, such as occlusion, changes in illumination, and facial expressions. As a result, improving the robustness of algorithms, especially against occlusion, remains a central concern in this field.

The Sparse Representation-based Classification (SRC) model, proposed by Wright et al. \cite{wright2008robust}, significantly improves robustness to occlusion by representing a test sample as a sparse linear combination of training samples and modeling occlusion and noise as a sparse error term. Naseem et al. \cite{naseem2012robust} adopted a block-wise and evidence fusion strategy to further enhance the robustness to contiguous occlusion and reduce the computational cost of SRC. Zhang et al. \cite{zhang2011sparse} pointed out that the key to face image classification lies in collaborative representation rather than sparse constraint, and thus proposed collaborative representation-based classification (CRC) with $\ell_2$-norm constraint. Cai et al. \cite{cai2016probabilistic} further proposed a robust collaborative representation-based classifier (RCRC), which adaptively suppresses the error influence of noise and occlusion points through an iterative reweighting strategy, significantly enhancing the model's robustness while maintaining the efficiency of CRC. Li et al. \cite{li2019sparsity} proposed a sparsity augmented weighted collaborative representation based classification (SA-WCRC) method for image classification. Yang et al. \cite{yang2011robust} proposed robust sparse coding (RSC), which constrains the error with the $\ell_1$-norm and adaptively assigns weights to different pixels during the sparse coding process through an iterative reweighting mechanism, effectively mitigating the influence of occluded regions and further improving recognition performance. Qian et al. \cite{qian2015robust} proposed a robust nuclear norm regularized regression (RNR) method for face recognition. Yang et al. \cite{yang2016nuclear} proposed a nuclear norm-based matrix regression (NMR) method, which models the error as a 2D image matrix, avoiding the loss of structural information caused by vectorization in traditional methods. Xie et al. \cite{xie2017robust} proposed a robust nuclear norm-based matrix regression model (RMR), which uses a non-convex function to characterize the low-rank structure of the error image. Chen et al. \cite{2019A} replaced the $\ell_2$-norm constraint in the NMR framework with the $\ell_1$-norm, proposing a sparse regularized NMR (SR\_NMR), and used training samples to learn a linear classifier for efficient classification. Li et al. \cite{2022Enhanced} proposed an enhanced nuclear norm-based matrix regression (ENMR) model, which imposes nuclear norm constraints on both the representation residual and the reconstructed image. Liu et al. \cite{2024A} proposed a robust face recognition model based on logarithmic non-convex relaxation regularization regression (log-NCRR), using a weighted non-convex approximation of the $\ell_{2,1}$-norm to ensure group sparsity of regression coefficients.

In addition to advanced classification algorithms, the effectiveness of a robust face recognition system also depends on powerful feature extraction. Compared with descriptors based on pixel intensity or local texture, gradient features can more effectively characterize the structural information of an image and have gradually received attention in face recognition. Dalal et al. \cite{2005Histograms} proposed histograms of oriented gradients (HOG), which constitute features by computing and stating the gradient direction histograms of local image regions, and applied it to human detection. Albiol et al. \cite{2008Face} successfully applied HOG descriptors to face recognition problems. O. et al. \cite{O2011Face} followed the ideas of Albiol et al., further exploring the representational capability of HOG features in face recognition and proposed a better method to construct robust HOG descriptors. Qian et al. \cite{2020Image} proposed a method called ID-NMR, which decomposes an image into multiple gradient images using local gradient distribution. Zhang et al. \cite{jianxin2020local} proposed a local occluded face recognition method based on block-based HOG and local binary pattern (LBP) features and sparse representation, effectively improving the robustness of face recognition. Chen et al. \cite{2021A} proposed a fusion algorithm based on LBP and HOG, which has stronger robustness under complex illumination conditions. Wu et al. \cite{2021Multispectral} proposed a maximum gradient and edge orientation histogram (HGEO) for multispectral image matching.

However, existing HOG-based methods focus solely on first-order gradient information, neglecting the valuable information contained in second- and higher-order gradients. Recent research in visual science suggests that neural images can be interpreted as a “surface” or “landscape,” whose geometric properties can be characterized by local curvature in differential geometry—properties that are precisely revealed by second-order gradient information \cite{2014HSOG,2011Features}. Based on second-order gradients, Huang et al. \cite{2014HSOG} proposed Histograms of second-order gradients (HSOG) to capture local geometric features related to curvature. Li et al. \cite{2019Defect} introduced a patterned fabric defect detection method using second-order direction-aware descriptors. Bastian et al. \cite{2019Pedestrian} developed a pedestrian detector that integrates first- and second-order Aggregate Channel Features. Zhang et al. \cite{2020No} designed a blind image quality assessment (IQA) method based on multi-order gradient statistics. Ye et al. \cite{ye2022robust} constructed a novel structural descriptor, the first- and second-order steerable filters (SFOC), which combines first- and second-order gradient information using steerable filters combined with a multi-scale strategy to capture more discriminative structural features. Yin et al. \cite{yin2022face} proposed a face recognition method based on compact second-order image gradient orientations, effectively improving robustness against real disguises, synthetic occlusions, and mixed variations.

Although methods such as HSOG have been effective, they often rely on complex feature construction techniques (e.g., DAISY-style ring pooling) to enhance discriminability, which inevitably increases computational cost. Inspired by this, we first propose a histogram of oriented hessian (HOH) directly based on the Hessian matrix, which effectively captures local geometric properties related to curvature by statistically aggregating the principal direction and magnitude of second-order gradients over a uniform grid. Furthermore, HOG and HOH features are computed in parallel on the same cell grid and combined at the block level with normalization, resulting in an efficient hybrid second-order gradient histogram (H2H) feature descriptor. This descriptor retains the complementary advantages of first- and second-order gradients while simplifying the feature construction process and improving computational efficiency. To further enhance the model's robustness to structured noise and large-area occlusion, the H2H features are incorporated into the SR\_NMR framework, and the residual modeling process is augmented with a global low-rank constraint. This integration gives rise to the hybrid second-order gradient histogram based global low-rank sparse regression (H2H-GLRSR) model. In contrast to existing methods that impose sample-wise low-rank constraints on the residuals (e.g., sample-wise nuclear norm minimization), our model imposes a global nuclear norm constraint, applying a joint low-rank regularizer to the entire residual matrix, thereby effectively capturing the global correlations among residuals and improving the face recognition performance in complex environments.

The main contributions of the paper are summarized as follows.

1. This paper proposes the H2H descriptor. HOH features are first constructed based on the Hessian matrix to capture the second-order geometric properties of images, such as curvature. Within the same cell-block framework, these HOH features are concatenated and normalized with HOG features to form a powerful H2H descriptor.

2. We integrate the H2H descriptor with SR$\_$NMR and introduce a global low-rank constraint on the residual matrix, resulting in the H2H-GLRSR model, which effectively handles structured noise and occlusion from a global perspective.

3. A fast optimization algorithm is used to solve the H2H-GLRSR model, and its computational complexity is analyzed. Experimental results demonstrate that the proposed method outperforms representative robust regression-based classification approaches on public face datasets.

The rest of this paper is organized as follows: Section 2 reviews traditional HOG features and the SR$\_$NMR algorithm; Section 3 introduces the proposed H2H-GLRSR model and the fast optimization algorithm for solving it; Section 4 presents experimental results and comparative analysis on public face datasets; Section 5 concludes the paper. 

\section{Related Work}
In this section, we provide a review of the traditional HOG descriptor and the fundamental principles of SR$\_$NMR.

\subsection{Histogram of oriented gradients (HOG)}
HOG, proposed by Dalal and Triggs \cite{2005Histograms}, is one of the most widely used feature descriptors for object detection and recognition tasks. The central idea of HOG is to represent the shape and contour of an object by analyzing the distribution of gradient orientations within local image regions.

Given a grayscale image $I \in \mathbb{R}^{p \times q}$, the calculation process of the HOG feature descriptor is as follows. First, the first-order gradient field of the image is computed using one-dimensional central difference filters $K_x = [-1, 0, 1]$ and $K_y = K_x^T$. These kernels serve as horizontal and vertical derivative operators, respectively. After convolution, the horizontal gradient map $I_x$ and the vertical gradient map $I_y$ are obtained as:
\begin{equation}
I_x = I * K_x, \quad I_y = I * K_y,
\end{equation}
where the notation ``$*$'' is the convolution operator. Based on $I_x$ and $I_y$, the gradient orientation $\theta(x, y)$ and magnitude $M(x, y)$ of each pixel are computed as:
\begin{equation}
\theta(x, y) = \arctan2(I_y(x,y), I_x(x,y)),
\end{equation}
\begin{equation}
M(x, y) = \sqrt{I_x(x,y)^2 + I_y(x,y)^2},
\end{equation}
where $\arctan2$ ensures correct angle computation in the range $(-\pi, \pi]$ by considering the signs of both $I_x$ and $I_y$.

Next, the image is divided into small cells of size $N \times N$ (e.g., $8 \times 8$). Within each cell, gradient orientations are quantized into $B$ discrete bins, usually spanning either $0$-$180^\circ$ (unsigned gradients) or $0$-$360^\circ$ (signed gradients). Each pixel casts a weighted vote for its corresponding orientation bin according to its gradient magnitude. The resulting histogram for each cell forms a $B$-dimensional vector, representing local edge information.

\subsection{Sparse regularized nuclear norm based matrix regression (SR\_NMR)}
Given a set of training images matrices \( \pmb{A}_1, \pmb{A}_2, \cdots,\pmb{A}_n \in \mathbb{R}^{p \times q} \) and a test image matrix \( \pmb{B} \in \mathbb{R}^{p \times q} \). The test image can be expressed as a linear combination of training samples as follows:
\begin{equation}
\pmb{B} = x_1 \pmb{A}_1 + x_2 \pmb{A}_2 + \cdots + x_n \pmb{A}_n + \pmb{E},
\end{equation}
or equivalently,
\begin{equation}
 \pmb{B} = A(\pmb{x}) + \pmb{E},
\end{equation}
where \( \pmb{x} = [x_1, x_2, \cdots, x_n]^{T} \in \mathbb{R}^{n} \) is the representation coefficient vector, \( \pmb{E} \in \mathbb{R}^{p \times q} \) is the representation error.

When the test image matrix $\pmb{B}$ is disturbed by contiguous occlusion, NMR assumes that the error matrix $\pmb{E}$ approximately exhibits low-rank structure. Under this assumption, the regression coefficients $\pmb{x}$ can be estimated by solving the following optimization problem:
\begin{equation}
\min_{\pmb{x}, \pmb{E}} \|\pmb{E}\|_{*} + \frac{1}{2} \lambda \|\pmb{x}\|_{2}^{2}, \quad \text{s.t.}\quad \pmb{B} = {A}(\pmb{x}) + \pmb{E}.
\end{equation}

Chen et al. \cite{2019A} argued that imposing an $\ell_1$-norm constraint on $\pmb{x}$ yields more discriminative and sparse representations than the $\ell_2$-norm. Accordingly, the SR\_NMR model replaces the $\ell_2$-norm with the $\ell_1$-norm, leading to:
\begin{equation}
\min_{\pmb{x}, \pmb{E}} \|\pmb{E}\|_{*} + \lambda \|\pmb{x}\|_{1}, \quad \text{s.t.}\quad \pmb{B} = {A}(\pmb{x}) + \pmb{E}.
\end{equation}

\section{The proposed method}

In this section, we first introduce the proposed HOH, which captures second-order geometric characteristics such as curvature and ridge-like structures. We then develop the H2H by integrating HOH with first-order gradient features, forming a unified representation of local structural information. Subsequently, the H2H features are incorporated into the SR\_NMR model. By imposing a global low-rank constraint on the residual matrix, the resulting model, referred to as H2H-GLRSR, effectively exploits cross-sample correlations in structured noise while maintaining sparsity. Furthermore, we use an efficient ADMM-based optimization algorithm for solving the model, followed by a theoretical complexity analysis and a ridge regression–based classification strategy.


\subsection{Histogram of oriented hessian (HOH)}
The HOH descriptor extends the philosophy of HOG by incorporating the second-order differential information contained in the Hessian matrix, thereby enhancing its ability to represent local geometric variations.

Given a grayscale image $I \in \mathbb{R}^{p \times q}$, the second-order derivatives are computed using convolution with second-order differential kernels:
\begin{equation}
I_{xx} = I * K_{xx}, \quad I_{yy} = I * K_{yy},\quad I_{xy} = I * K_{xy},
\end{equation}
where $K_{xx}$, $K_{yy}$ and $K_{xy}$ are the corresponding discrete convolution kernels.

For each pixel point $(x, y)$ in the image, its $Hessian$ matrix is defined as:
\begin{equation}
\pmb{H}(x,y)=\begin{bmatrix}
I_{xx}(x,y) & I_{xy}(x,y) \\
I_{xy}(x,y) & I_{yy}(x,y)
\end{bmatrix}.
\end{equation}
From this matrix, the principal curvature orientation $\phi(x, y)$ and corresponding gradient strength $S(x, y)$ are derived analytically as:
\begin{equation}
\phi(x,y) = \frac{1}{2} \arctan 2\left( 2I_{xy}(x,y),\, I_{xx}(x,y) - I_{yy}(x,y)\right),
\end{equation}
\begin{equation}
S(x,y)=\sqrt{(I_{xx}(x,y)-I_{yy}(x,y))^2+4I_{xy}^2(x,y)}.
\end{equation}

Each image cell accumulates a weighted histogram of the orientations $\phi$ of its pixels, where the weights correspond to $S(x,y)$. The resulting histogram forms the HOH descriptor for that cell.

Compared with HOG, HOH is highly sensitive to second-order image variations and effectively complements the contour information captured by first-order gradients, thereby forming a more comprehensive and more discriminative representation of local structural features.

\subsection{Hybrid second-order gradient histogram (H2H)}
The H2H descriptor combines HOG and HOH in a unified computational framework to achieve both discriminative power and computational efficiency. The construction of H2H involves three major steps:

(1) Parallel feature extraction.

HOG and HOH features are computed in parallel on the same grid of cells. For each cell, the histogram bins are calculated as:
\begin{equation}
HOG(b) = \sum_{(x,y) \in \text{cell}} M(x,y) \cdot \mathbf{1}\{\theta(x,y) \in \text{bin} \, b\},
\end{equation}

\begin{equation}
HOH(b) = \sum_{(x,y) \in \text{cell}} S(x,y) \cdot \mathbf{1}\{\phi(x,y) \in \text{bin} \, b\},
\end{equation}
where $\mathbf{1}\{\cdot\}$ is the indicator function, which is 1 when the condition is true and 0 otherwise; $\text{bin} \, b$ denotes the $b$-th orientation bin, $b = 1,2,...,B$, and $B$ is the total number of bins.

Thus, each cell obtains two $B$-dimensional row vectors:
\begin{equation}
\pmb{h}_{hog} = [HOG(1), HOG(2), \dots, HOG(B)],
\end{equation}
\begin{equation}
\pmb{h}_{hoh} = [HOH(1), HOH(2), \dots, HOH(B)].
\end{equation}
The two vectors are then jointly normalized using the $\ell_2$ norm:

\begin{equation}
\pmb{\hat{h}}_{\text{hog}} = \frac{\pmb{h}_{\text{hog}}}{\sqrt{\|\pmb{h}_{\text{hog}}\|_2^2 + \|\pmb{h}_{\text{hoh}}\|_2^2 }},
\end{equation}

\begin{equation}
\pmb{\hat{h}}_{\text{hoh}} = \frac{\pmb{h}_{\text{hoh}}}{\sqrt{\|\pmb{h}_{\text{hog}}\|_2^2 + \|\pmb{h}_{\text{hoh}}\|_2^2 }},
\end{equation}
where $\widehat{\pmb{h}}_{hog}$ and $ \widehat{\pmb{h}}_{hoh}$ are the normalized HOG and HOH feature vectors, respectively.

(2) Block feature construction and normalization.

Adjacent $2 \times 2$ cells are grouped to form a block, where each cell corresponds to a local region of size $N \times N$ pixels. Within each block, the normalized cell-level HOG feature vectors are concatenated in a row-major order, meaning that the vectors of the top-left, top-right, bottom-left, and bottom-right cells are arranged sequentially according to their spatial positions. Afterwards, the corresponding HOH vectors from the same four cells are concatenated in the same order to maintain spatial consistency between first- and second-order gradient features. This process produces a block-level feature vector:
\begin{equation}
\pmb{f}_{\text{block}} = [\pmb{\hat{h}}_{\text{hog}}^1, \pmb{\hat{h}}_{\text{hog}}^2, \pmb{\hat{h}}_{\text{hog}}^3, \pmb{\hat{h}}_{\text{hog}}^4, \pmb{\hat{h}}_{\text{hoh}}^1, \pmb{\hat{h}}_{\text{hoh}}^2, \pmb{\hat{h}}_{\text{hoh}}^3, \pmb{\hat{h}}_{\text{hoh}}^4],
\end{equation}
which is then $\ell_2$-norm normalized to ensure illumination invariance and feature balance:
\begin{equation}
\pmb{\hat{\mathbf{f}}}_{\text{block}} = \frac{\pmb{\mathbf{f}}_{\text{block}}}{\|\pmb{\mathbf{f}}_{\text{block}}\|_2}.
\end{equation}
Therefore, each block feature has a dimensionality of $8B$.

(3) Global feature concatenation.

All possible block positions are traversed using a sliding window with stride one (one cell). The total number of blocks is:
\begin{equation}
K = (N_y - 1) \times (N_x - 1) \tag{16},
\end{equation}
where \(N_y\) and \(N_x\) are denote the number of cells along the vertical and horizontal directions, respectively. The final H2H descriptor for the image is obtained by concatenating all normalized block features in sequence:
\begin{equation}
\pmb{f} = [\hat{\mathbf{f}}_{\text{block}}^1, \hat{\mathbf{f}}_{\text{block}}^2, \ldots, \hat{\mathbf{f}}_{\text{block}}^K]^T \in \mathbb{R}^{8BK}.
\end{equation}

\begin{figure}[htbp]
    \centering
    \begin{subfigure}[c]{0.33\textwidth}
        \centering
        \includegraphics[width=\linewidth]{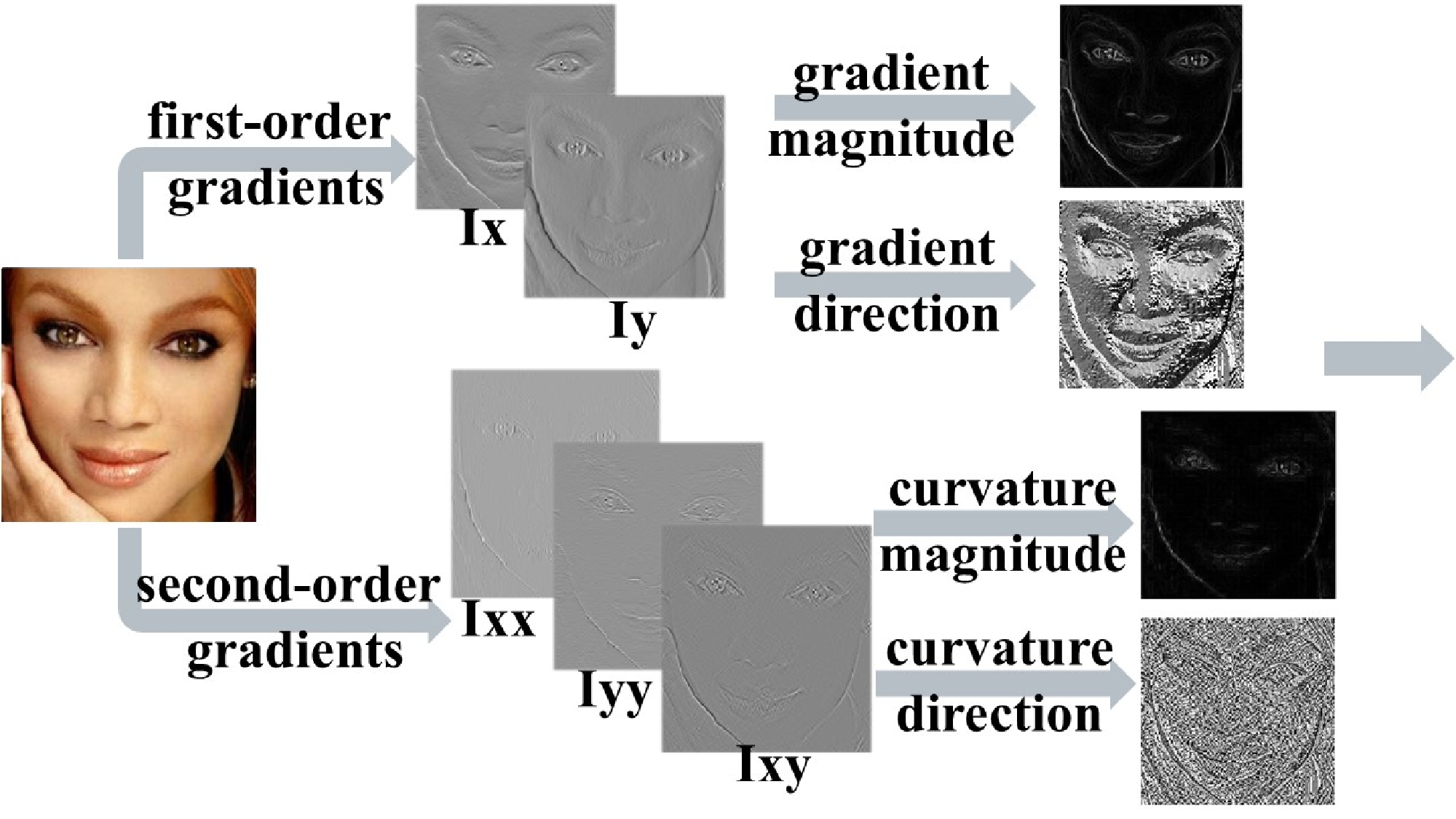}
    \end{subfigure}
    \hfill
    \begin{subfigure}[c]{0.33\textwidth}
        \centering
        \includegraphics[width=\linewidth]{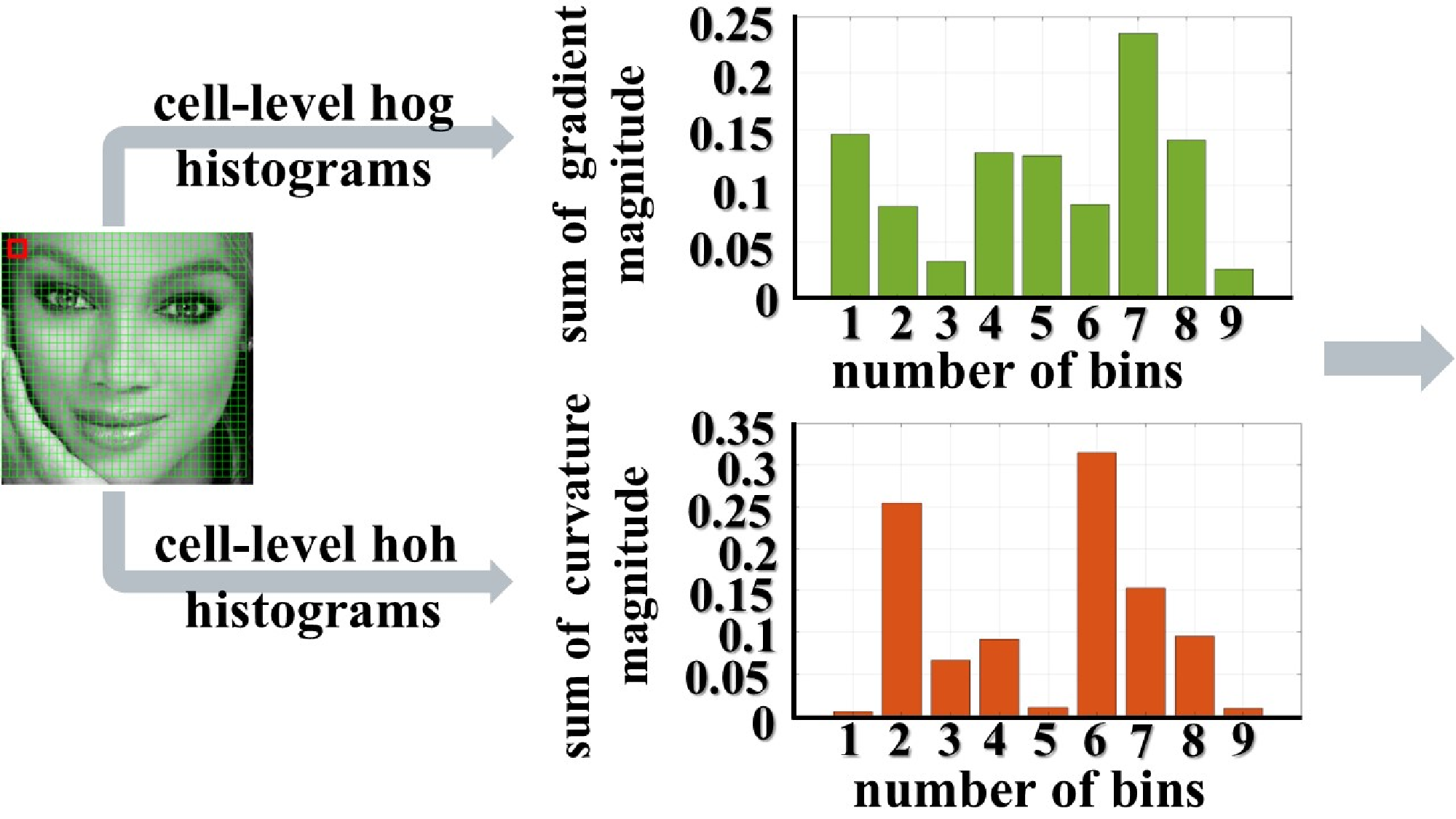}
    \end{subfigure}
    \hfill
    \begin{subfigure}[c]{0.3\textwidth}
        \centering       \includegraphics[height=2.6cm,valign=b]{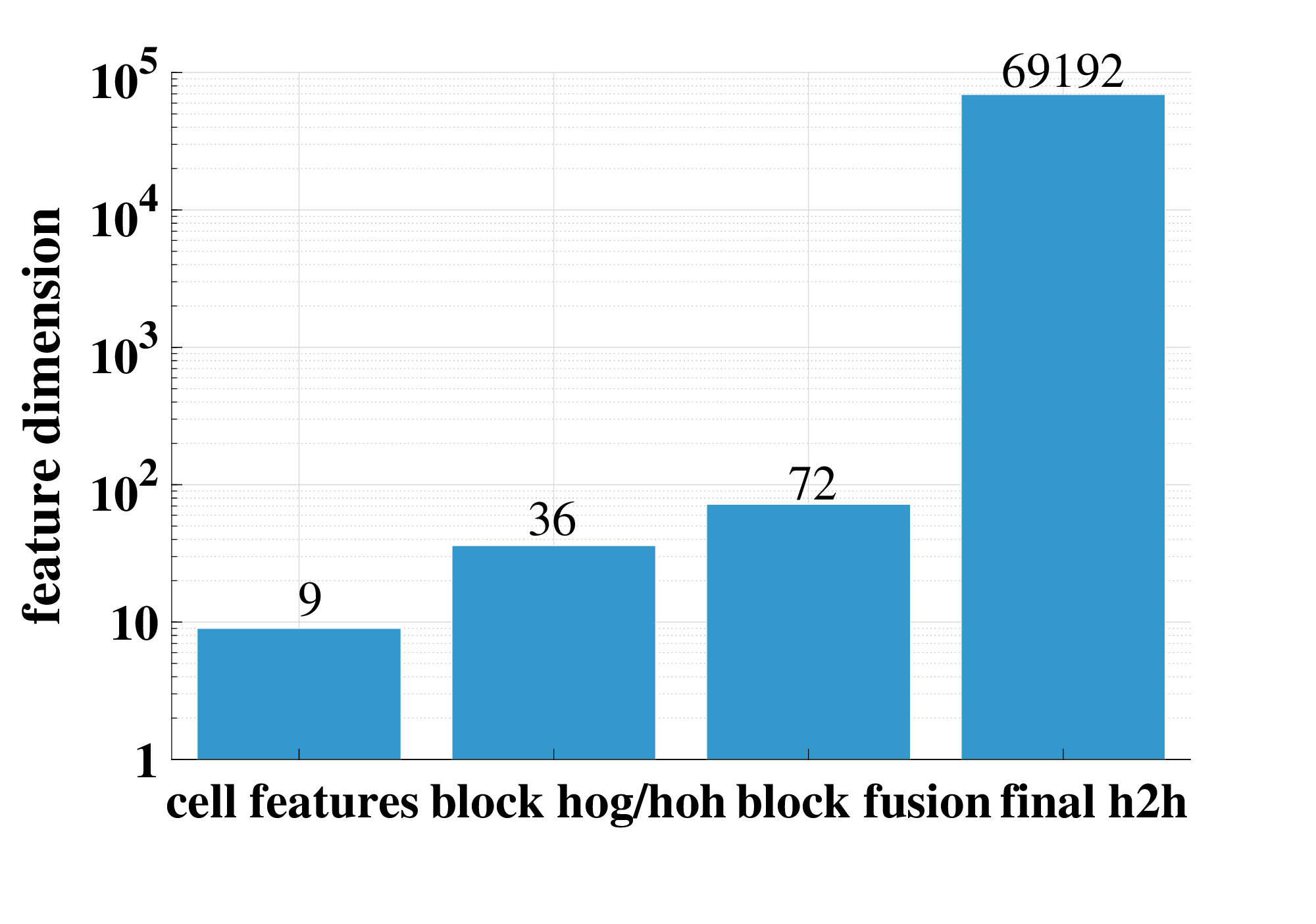}
    \end{subfigure}
    \caption{Flowchart for constructing H2H feature descriptors}
    \label{fig:flowchart for Constructing H2H Feature Descriptors}
\end{figure}

The proposed H2H descriptor retains the respective advantages of HOG in edge description and HOH in curvature characterization. Moreover, its two-level normalization process improves feature stability and discriminability, making H2H both efficient and robust for visual representation tasks. The construction flowchart of the H2H descriptor is illustrated in Fig.~\ref{fig:flowchart for Constructing H2H Feature Descriptors}.

\subsection{Hybrid second-order gradient histograms based global low-rank sparse regression (H2H-GLRSR)}

To overcome the limitations of SR\_NMR in handling structured noise and correlated occlusions across samples, a global low-rank constraint is imposed on the residual matrix. Incorporating this constraint with the H2H features gives rise to the H2H-GLRSR model, which simultaneously leverages the discriminative power of H2H features and captures the global correlation structure of the residuals.


Given a set of training images matrices \( \pmb{A}_1, \pmb{A}_2, \cdots,\pmb{A}_n \in \mathbb{R}^{p \times q} \) and a test image matrix \( \pmb{B} \in \mathbb{R}^{p \times q} \), each image $\pmb{A}_i$ can be mapped to a high-dimensional feature vector $\Phi(\pmb{A}_i) \in \mathbb{R}^d$ via the H2H descriptor, where $\Phi(\cdot)$ denotes the H2H feature extraction operator and $d$ is the feature dimension. Combine the H2H feature vectors of all training images to obtain the training sample matrix:
\begin{equation}
 \pmb{X}_{tr} = [\Phi(\pmb{A}_1), \Phi(\pmb{A}_2), \cdots, \Phi(\pmb{A}_n)] \in \mathbb{R}^{d \times n}.
\end{equation}
Similarly, extract the H2H feature for the test image $\pmb{B}$, denoted as:
\begin{equation}
\pmb{y} = \Phi(\pmb{B}) \in \mathbb{R}^d.
\end{equation}

Under the traditional SR\_NMR framework, the test feature $\pmb{y}$ can be linearly represented by:
\begin{equation}
\pmb{y} = \pmb{X}_{tr} \pmb{z} + \pmb{e},
\end{equation}
where $\pmb{z} \in \mathbb{R}^n$ is the regression coefficient vector and $\pmb{e} \in \mathbb{R}^d$ is the residual. Imposing the $\ell_1$ sparse constraint on $\pmb{z}$ ensures that the test sample is mainly represented by training samples of the same class.

However, when multiple test samples are affected by structured noise or occlusions, treating their residuals independently fails to capture the global dependencies among them. To address this issue, we jointly model all residuals as a residual matrix $\pmb{E} = [\pmb{e}_1, \pmb{e}_2, \cdots, \pmb{e}_m] \in \mathbb{R}^{d \times m}$,and impose a nuclear norm regularization term on $\pmb{E}$ to enforce a low-rank structure. The resulting optimization problem is formulated as:

\begin{equation}\label{eq:H2H-GLRSR}
\min_{\pmb{Z}, \pmb{E}} \lambda \| \pmb{Z} \|_1 + \alpha \| \pmb{E} \|_*, \quad \text{s.t.} \quad \pmb{Y} = \pmb{X}_{tr} \pmb{Z} + \pmb{E},
\end{equation}
where:
\begin{itemize}
\item $\pmb{Y} = [\pmb{y}_1, \pmb{y}_2, \cdots, \pmb{y}_m] \in \mathbb{R}^{d \times m}$ is the H2H feature matrix of test samples;
\item $\pmb{Z} = [\pmb{z}_1, \pmb{z}_2, \cdots, \pmb{z}_m] \in \mathbb{R}^{n \times m}$ is the regression coefficient matrix;
\item $\lambda, \alpha > 0$ are regularization parameters controlling sparsity and low-rank strength, respectively.
\end{itemize}

\subsubsection{Fast optimization algorithm for H2H-GLRSR}
To efficiently solve the optimization problem \eqref{eq:H2H-GLRSR}, the ADMM \cite{boyd2011distributed} is employed. First, introduce an auxiliary variable $\pmb{J}$ and rewrite formula \eqref{eq:H2H-GLRSR} as:
\begin{equation}\label{eq:new_H2H-GLRSR}
\min_{\pmb{Z}, \pmb{J}, \pmb{E}} \lambda \| \pmb{J} \|_1 + \alpha \| \pmb{E} \|_*, \quad \text{s.t.} \quad
\pmb{Y} = \pmb{X}_{tr} \pmb{Z} + \pmb{E},\quad \pmb{Z} = \pmb{J}.
\end{equation}
The augmented Lagrangian formulation of \eqref{eq:new_H2H-GLRSR} is formulated as follows:
\begin{align}\label{eq:Lagrangian function}
\min_{\pmb{Z}, \pmb{J}, \pmb{E}}
&\lambda \| \pmb{J} \|_1 + \alpha \| \pmb{E} \|_* \nonumber+ \langle \pmb{T}_1, \pmb{Y} - \pmb{X}_{tr} \pmb{Z} - \pmb{E} \rangle + \langle \pmb{T}_2, \pmb{Z} - \pmb{J} \rangle \nonumber\\
& + \frac{\mu}{2} \| \pmb{Y} - \pmb{X}_{tr} \pmb{Z} - \pmb{E} \|_F^2 + \frac{\mu}{2} \|\pmb{Z} - \pmb{J} \|_F^2,
\end{align}
where $\pmb{T}_1$ and $\pmb{T}_2$ are Lagrange multipliers, and $\mu > 0$ is a penalty parameter. 


(1) Updating $\pmb{Z}$. After dropping the irrelevant terms with respect to $\pmb{Z}$,
the objective function of model \eqref{eq:Lagrangian function} can be written as
\begin{equation}\label{eq: the subproblem of Z}
\min_{\pmb{Z}} \frac{\mu}{2} \left\| \pmb{Y} - \pmb{X}_{tr} \pmb{Z} - \pmb{E} + \frac{1}{\mu} \pmb{T}_1 \right\|_F^2 + \frac{\mu}{2} \left\| \pmb{Z}- \pmb{J} + \frac{1}{\mu} \pmb{T}_2 \right\|_F^2.
\end{equation}
The closed-form solution of \eqref{eq: the subproblem of Z} is
\begin{equation}\label{eq:Z}
\pmb{Z} ^{k+1} = \pmb{C} \left[\pmb{X}_{tr}^{\top} \left(\pmb{Y}- \pmb{E}^{k} + \frac{1}{\mu^{k}} \pmb{T}_1^{k} \right) + \pmb{J}^{k} - \frac{1}{\mu^{k}} \pmb{T}_2^{k} \right],
\end{equation}
where $\pmb{C} = (\pmb{X}_{tr}^{\top} \pmb{X}_{tr} + \pmb{I}_n)^{-1}$.

(2) Updating $\pmb{J}$. By ignoring terms independent of $\pmb{J}$, 
the problem \eqref{eq:Lagrangian function} can be reformulated as follows:
\begin{equation}\label{eq: the subproblem for J}
\min_{J} \lambda \| \pmb{J} \|_1 + \frac{\mu}{2} \left\| \pmb{Z} - \pmb{J} + \frac{1}{\mu} \pmb{T}_2 \right\|_F^2.
\end{equation}

Problem \eqref{eq: the subproblem for J} can be solved by using the element wise soft-thresholding, then we get

\begin{equation}\label{eq:J}
\pmb{J}^{k+1} = S_{\lambda / \mu^{k}} \left( \pmb{Z}^{k+1} + \frac{1}{\mu^{k}} \pmb{T}_2^{k} \right),
\end{equation}
where \( S_{\lambda / \mu^{k}}(\cdot) \) is the soft-thresholding operator \cite{lin2010augmented}, which is defined as
\begin{equation}
S_{\xi}(\pi) = 
\begin{cases} 
\pi - \xi, & \text{if } \pi > \xi \\ 
\pi + \xi, & \text{if } \pi < -\xi \\ 
0, & \text{otherwise}
\end{cases}
\end{equation}

(3) Updating $\pmb{E}$. After removing the terms unrelated to $\pmb{E}$, \eqref{eq:Lagrangian function} can be represented as
\begin{equation}\label{eq:the subproblem for E}
\min_{\pmb{E}} \alpha \|\pmb{E}\|_* + \frac{\mu}{2} \left\| \pmb{Y} - \pmb{X}_{tr} \pmb{Z} - \pmb{E} + \frac{1}{\mu} \pmb{T}_1 \right\|_F^2.
\end{equation}

By utilizing the singular value thresholding operator algorithm \cite{2010A}, problem \eqref{eq:the subproblem for E} has a closed-from solution:
\begin{equation}\label{eq:E}
\pmb{E}^{k+1} = T_{\alpha / \mu^{k}}(\pmb{W}) = \pmb{U} S_{\alpha / \mu^{k}}(\pmb\Sigma) \pmb{V}^\top,
\end{equation}
where \( \pmb{W} = \pmb{Y} - \pmb{X}_{tr} \pmb{Z}^{k+1} + \frac{1}{\mu^{k}} \pmb{T}_1^{k} \), and \( \pmb{W} = \pmb{U} \pmb\Sigma \pmb{V}^\top \) is the singular value decomposition of $\pmb{W}$. 
The complete updating procedure is presented in Algorithm \ref{alg:Optimization Procedure of H2H-GLRSR}. 

\begin{algorithm}[htbp]
\caption{Optimization Procedure of H2H-GLRSR} \label{alg:Optimization Procedure of H2H-GLRSR}
\begin{algorithmic}
\State \textbf{Input:} Training samples H2H feature matrix \(\pmb{X}_{tr} \in \mathbb{R}^{d \times n}\), test samples H2H feature matrix \(\pmb{Y} \in \mathbb{R}^{d \times m}\), parameters \(\lambda, \alpha\).
\State \textbf{Initialization:} \(\pmb{Z}= 0_{n \times m},\, \pmb{J} = 0_{n \times m}, \, \pmb{E} = 0_{d \times m}, \, \pmb{T}_1 = 0_{d \times m},\,
\pmb{T}_2 = 0_{n \times m}, \, \mu_{\max} = 10^{10}, \, \text{tol} = 10^{-6}, \, \rho = 1.5\).
\State 1. Compute \(\pmb{C} = (\pmb{X}_{tr}^{\top} \pmb{X}_{tr} + \pmb{I}_n)^{-1}\);
\While{not converged do}
\State 2. Update \(\pmb{Z}\) by using formula \eqref{eq:Z};
\State 3. Update \(\pmb{J}\) by using formula \eqref{eq:J};
\State 4. Update \(\pmb{E}\) by using formula \eqref{eq:E};
\State 5. Update Lagrange multipliers \(\pmb{T}_1\) and \(\pmb{T}_2\);
\State \quad 1) Update \(\pmb{T}_1: \pmb{T}_1^{k+1} = \pmb{T}_1^k + \mu^k(\pmb{Y} - \pmb{X}_{tr}\pmb{Z}^{k+1} - \pmb{E}^{k+1})\)
\State \quad 2) Update \(\pmb{T}_2: \pmb{T}_2^{k+1} = \pmb{T}_2^k + \mu^k(\pmb{Z}^{k+1} - \pmb{J}^{k+1})\)
\State 6. Update \(\mu: \mu^{k+1} = \min(\mu_{\max}, \, \rho\mu^k)\);
\State 7. Check convergence:
\State \quad \(\max\left\{\|\pmb{Y} - \pmb{X}_{tr}\pmb{Z}^{k+1} - \pmb{E}^{k+1}\|_{\infty}, \|\pmb{Z}^{k+1} - \pmb{J}^{k+1}\|_{\infty}\right\} < \text{tol}\).
\EndWhile
\State \textbf{Output:} \(\pmb{Z}\)
\end{algorithmic}
\end{algorithm}


\subsection{Computational Complexity Analysis}

Similar to \cite{2019A}, we neglect the computational cost of simple matrix addition, subtraction, and scalar matrix operations, since they are negligible compared to matrix decomposition. The main computational burden of the proposed algorithm comes from two parts:  
(1) calculating the inverse of matrix in \eqref{eq:Z};  
(2) the low-rank reconstruction of the residual matrix $\pmb{E}$.

Before the iteration begins, the complexity of pre-computing $(\pmb{X}_{tr}^{\top} \pmb{X}_{tr} + \pmb{I}_n)^{-1}$
in \eqref{eq:Z} is $O(n^3)$. In the residual update stage, all test samples are unified into a single residual matrix $\pmb{E}$, and its low-rank structure is recovered via a single singular value decomposition (SVD), the complexity of this SVD step is $O\big(\min\{d^2 m, d m^2\}\big)$. 
Since only one SVD is required for all test samples in each iteration, our approach avoids the repetitive computations needed for performing SVD on each test image separately as described in \cite{2019A}.

Therefore, the overall computational complexity of the proposed method is approximately $O\Big(n^3 + t \times \min\{d^2 m, d m^2\}\Big)$,where $t$ is the number of iterations.  


\subsection{H2H-GLRSR based classifier}
To perform classification based on the learned representation, we adopt a ridge regression–based classifier inspired by the Discriminative Block-Diagonal Low-Rank Representation (BDLRR) method \cite{zhang2017discriminative}. This classifier effectively alleviates the issues of matrix singularity and overfitting in small-sample settings while maintaining computational efficiency.

Let \( \pmb{Z}_{tr} \in \mathbb{R}^{n \times n_{tr}} \) denotes the sparse representation coefficient matrix of training samples, and  \( \pmb{H}_{tr} \in \mathbb{R}^{c \times n_{tr}} \) be the corresponding one-hot encoded label matrix, where $c$ is the number of classes. The classifier learns the weight matrix \( \pmb{W} \in \mathbb{R}^{c \times n} \) by solving:
\begin{equation}\label{eq:W}
\min_{\pmb{W}} \|\pmb{W} \pmb{Z}_{tr} - \pmb{H}_{tr}\|_F^2 + \eta \|\pmb{W}\|_F^2,
\end{equation}
where \(\eta > 0\) is a regularization parameter. The closed-form solution of \eqref{eq:W} is:
\begin{equation}
\pmb{W} = \pmb{H}_{tr} \pmb{Z}_{tr}^T (\pmb{Z}_{tr} \pmb{Z}_{tr}^T + \eta \pmb{I})^{-1}.
\end{equation}
For stability, the weight matrix is normalized column-wise as:
\begin{equation}  
\boldsymbol{W} \leftarrow \boldsymbol{W} \cdot \operatorname{diag}\left(\frac{1}{\|\boldsymbol{W}_{:,j}\|_2}\right), 
\end{equation}
where $\|\boldsymbol{W}_{:,j}\|_2$ is the $\ell_{2}$-norm of the $j$-th column of $\boldsymbol{W}$.

Given a test representation coefficient matrix \( \pmb{Z}_{tt} \in \mathbb{R}^{n \times n_{tt}}\), the predicted class scores are obtained by:
\begin{equation}
\hat{\pmb{H}}_{tt} = \pmb{W} \pmb{Z}_{tt},
\end{equation}
The predicted class label for the $i$-th test sample is:
\begin{equation}
\hat{y}_i = \arg \max_j \hat{\pmb{H}}_{tt}(j, i), \quad i = 1, 2, \ldots, n_{tt}.
\end{equation}

\section{Experiments}

In this section, we evaluate the proposed method on three publicly available datasets, AR\cite{martinez1998ar}, Extended Yale B\cite{lee2005acquiring} and Labeled Faces in the Wild (LFW)\cite{huang2008labeled}. H2H-GLRSR is compared with other regression-based algorithms, including SRC\cite{wright2008robust}, RLRC\cite{naseem2012robust}, CRC\cite{zhang2011sparse}, RSC\cite{yang2011robust}, RNR\cite{qian2015robust}, NMR\cite{yang2016nuclear}, SR\_NMR\cite{2019A} and ID-NMR\cite{2020Image}. The parameter values of these algorithms are the same as those recommended by the original authors. Some of the experimental results are taken from the literature\cite{2020Image}. In our method, there are mainly five parameters: cell size \(N\), orientation bins \(B\) and regularization parameters $\lambda$, $\alpha$ and $\eta$. We set the classifier parameter $\eta$ is 1. The selection of the remaining parameters for different face recognition scenarios is detailed in Section 4.4. We focus on three realistic evaluation scenarios: real disguise (occlusion), illumination variation, and unconstrained face images.

\subsection{Face images with real disguise}
\begin{figure}[h]
\centering
\includegraphics[width=1\linewidth]{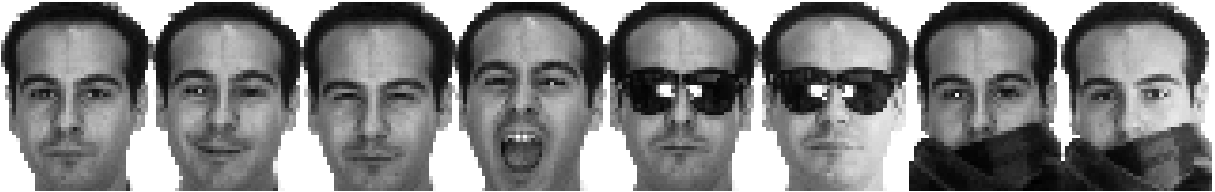}
\caption{Some sample images of one person from the AR database}\label{fig:AR}
\end{figure}

In this subsection, we evaluate the proposed method on the AR dataset. The AR database consists of about 4000 face images of 126 individuals (70 men and 56 women), among which 120 individuals can be divided into two sessions of equal size, each session containing 13 images. Following \cite{2019A}, we select 8 non-occluded images per person as training samples, and the remaining 6 images with scarf occlusion and 6 images with sunglasses occlusion are used as test set 1 and test set 2, respectively. Each image is cropped to $50 \times 40$ pixels. Fig.~\ref{fig:AR} shows some sample images of one person.

Table \ref{tab:AR_recognition_rates} summarizes recognition rates for each method under the two occlusion types. The results show that methods based on sparse or collaborative representations (SRC, CRC, RSC) retain high accuracy for local, structured occlusions such as sunglasses but degrade substantially for large, unstructured occlusions like scarves. Models that incorporate low-rank priors (RNR, NMR, SR\_NMR) improve performance on scarf occlusion, indicating the benefit of modeling structured error. The ID-NMR method further improves accuracy by decomposing images into gradient components. Our H2H-GLRSR method attains the highest recognition rates on both occlusion types (99.4\% for sunglasses and 96.1\% for scarves). We attribute this improvement primarily to two aspects: first, the construction of the H2H descriptor, which incorporates both first- and second-order gradient information; second, the imposition of a global low-rank constraint on the residual matrix in the regression model, which captures the correlations among occlusion errors across samples.

\begin{table}[h]
  \centering
  \caption{Recognition rates (\%) of each method for FR with real disguise on the AR Database.}
  \label{tab:AR_recognition_rates}
  \begin{tabular}{lcc}
    \toprule
    \textbf{Methods/cases} & \textbf{Sunglasses} & \textbf{Scarf} \\
    \midrule
    SRC\cite{wright2008robust} & 94.4            & 57.6           \\
    RLRC\cite{naseem2012robust}& 94.6            & 53.3           \\
    CRC\cite{zhang2011sparse}  & 93.5            & 63.3           \\
    RSC\cite{yang2011robust}   & 94.2            & 66.8           \\
    RNR\cite{qian2015robust}   & 97.2            & 76.8           \\
    NMR\cite{yang2016nuclear}  & 96.9            & 73.5           \\
    SR\_NMR\cite{2019A}        & 97.6            & 75.1           \\
    ID-NMR\cite{2020Image}     & 99.0            & 93.8           \\
    \textbf{H2H-GLRSR}        & \textbf{99.4}       & \textbf{96.1}  \\
    \bottomrule
  \end{tabular}
\end{table}

\subsection{Face images with illumination changes}
\begin{figure}[h]
\centering
\includegraphics[width=1\linewidth]{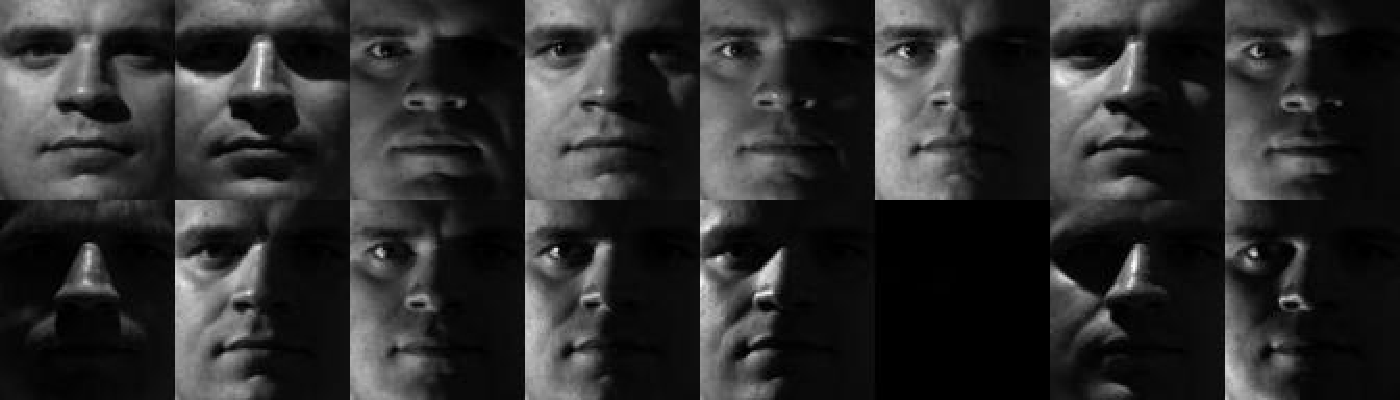}
\caption{Sample images with illumination changes collected from Subset 4 and Subset 5 of Extended Yale B. First row: Subset 4, Second row: Subset 5.}\label{fig:Extended Yale}
\end{figure}

In this subsection, we evaluate the proposed method on the Extended Yale B dataset. This dataset consists of 2414 face images of 38 individuals under different lighting conditions. Five subsets correspond to different degrees of illumination. We first select all images with natural illumination (Subset 1) as the training set, and images with extreme illumination changes (Subsets 4 and 5) are used as test set 1 and test set 2, respectively. Each image is cropped to $96 \times 84$ pixels. Fig.~\ref{fig:Extended Yale} shows some sample images with illumination changes.


Table \ref{tab:EYB_recognition_rates} reports the recognition accuracies of various methods under different illumination conditions on the Extended Yale B database. 
Traditional sparse or collaborative representation models (SRC, CRC, RSC) suffer dramatic performance degradation under extreme illumination (Subset 5), where their accuracies decrease to approximately 28.8\%\,--\,36.7\%, indicating limited robustness to lighting variations. More advanced low-rank or regularized models (RLRC, RNR, NMR, SR\_NMR) exhibit improved resistance to illumination but still experience noticeable drops when moving from Subset 4 to Subset 5.
Although the recently proposed ID-NMR achieves competitive results of 98.4\% and 99.4\% on the two subsets, respectively, H2H-GLRSR maintains superior performance on Subset 4 (99.8\%) and remains nearly comparable on Subset 5 (99.3\%), demonstrating strong overall robustness across illumination conditions.
The performance gain primarily stems from the introduction of HOH features, which capture curvature-related patterns and second-order geometric structures that are more resilient to drastic lighting changes. By exploiting the principal directions and magnitudes of second-order differentials, HOH enhances feature discriminability and stability. These results validate the effectiveness and unique advantages of incorporating second-order gradient information for illumination-robust face recognition.

\begin{table}[h]
  \centering
  \caption{Recognition rates (\%) of each method for FR under illumination changes on the Extended Yale B Database.}
  \label{tab:EYB_recognition_rates}
  \begin{tabular}{lcc}
    \toprule
    \textbf{Methods/cases} & \textbf{Subset 4} & \textbf{Subset 5} \\
    \midrule
    SRC\cite{wright2008robust} & 78.4            & 28.8           \\
    RLRC\cite{naseem2012robust}& 89.7            & 43.2           \\
    CRC\cite{zhang2011sparse}  & 88.0            & 35.7           \\
    RSC\cite{yang2011robust}   & 80.3            & 36.7           \\
    RNR\cite{qian2015robust}   & 91.6            & 49.0           \\
    NMR\cite{yang2016nuclear}  & 90.2            & 47.9           \\
    SR\_NMR\cite{2019A}        & 76.8            & 35.7           \\
    ID-NMR\cite{2020Image}     & 98.4            & \textbf{99.4}           \\
    \textbf{H2H-GLRSR}        & \textbf{99.8}       & 99.3  \\
    \bottomrule
  \end{tabular}
\end{table}

\subsection{Face images in unconstrained environment}
\begin{figure}[h]
\centering
\includegraphics[width=1\linewidth]{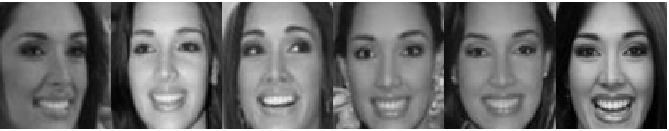}
\caption{Some sample images of one person from the LFW database}\label{fig:LFW}
\end{figure}

In this subsection, we mainly evaluate the proposed method on the LFW dataset. In this experiment, we use the aligned version of LFW, selecting face images of 100 subjects, each with 6 images. The first 4 images are used as the training set, and the remaining 2 images are used as the test set. Each image is cropped to $80 \times 70$ pixels. Fig.~\ref{fig:LFW} shows some sample images of one person.

Table \ref{tab:LFW_recognition_rates} reports the recognition performance of different methods on the unconstrained LFW face database. Traditional sparse representation methods (SRC, CRC) and robust regression methods (RSC, RLRC) are constrained by limited feature representation and insufficient noise suppression capabilities, leading to relatively modest performance. Low-rank constraint-based models (RNR, NMR, SR\_NMR) achieve moderate improvements but still treat sample residuals independently, failing to fully exploit the structural dependencies among errors.The ID-NMR method achieves notable improvement through enhanced feature representation, yet its performance remains constrained by its local regularization mechanism. 
In contrast, the H2H-GLRSR method achieves the optimal recognition accuracy of 74.5\% by fusing second-order gradient features and effectively suppressing structural noise through a global low-rank constraint.
This result verifies the robustness and superiority of the proposed method in complex, unconstrained environments.

\begin{table}[h]
  \centering
  \caption{Recognition rates (\%) of each method for FR in unconstrained environment on the LFW Database}
  \label{tab:LFW_recognition_rates}
  \begin{tabular}{lc}
    \toprule
    \textbf{Methods} & \textbf{LFW}  \\
    \midrule
    SRC\cite{wright2008robust} & 40.0                           \\
    RLRC\cite{naseem2012robust}& 30.6                           \\
    CRC\cite{zhang2011sparse}  & 43.0                             \\
    RSC\cite{yang2011robust}   & 41.6                             \\
    RNR\cite{qian2015robust}   & 46.6                             \\
    NMR\cite{yang2016nuclear}  & 44.5                             \\
    SR\_NMR\cite{2019A}        & 39.5                             \\
    ID-NMR\cite{2020Image}     & 62.5                             \\
    \textbf{H2H-GLRSR}        & \textbf{74.5}                    \\
    \bottomrule
  \end{tabular}
\end{table}

\subsection{Parameter selection}
\begin{figure}[htbp]
    \centering
    \begin{subfigure}{0.48\textwidth}
        \centering
        \includegraphics[width=\linewidth]{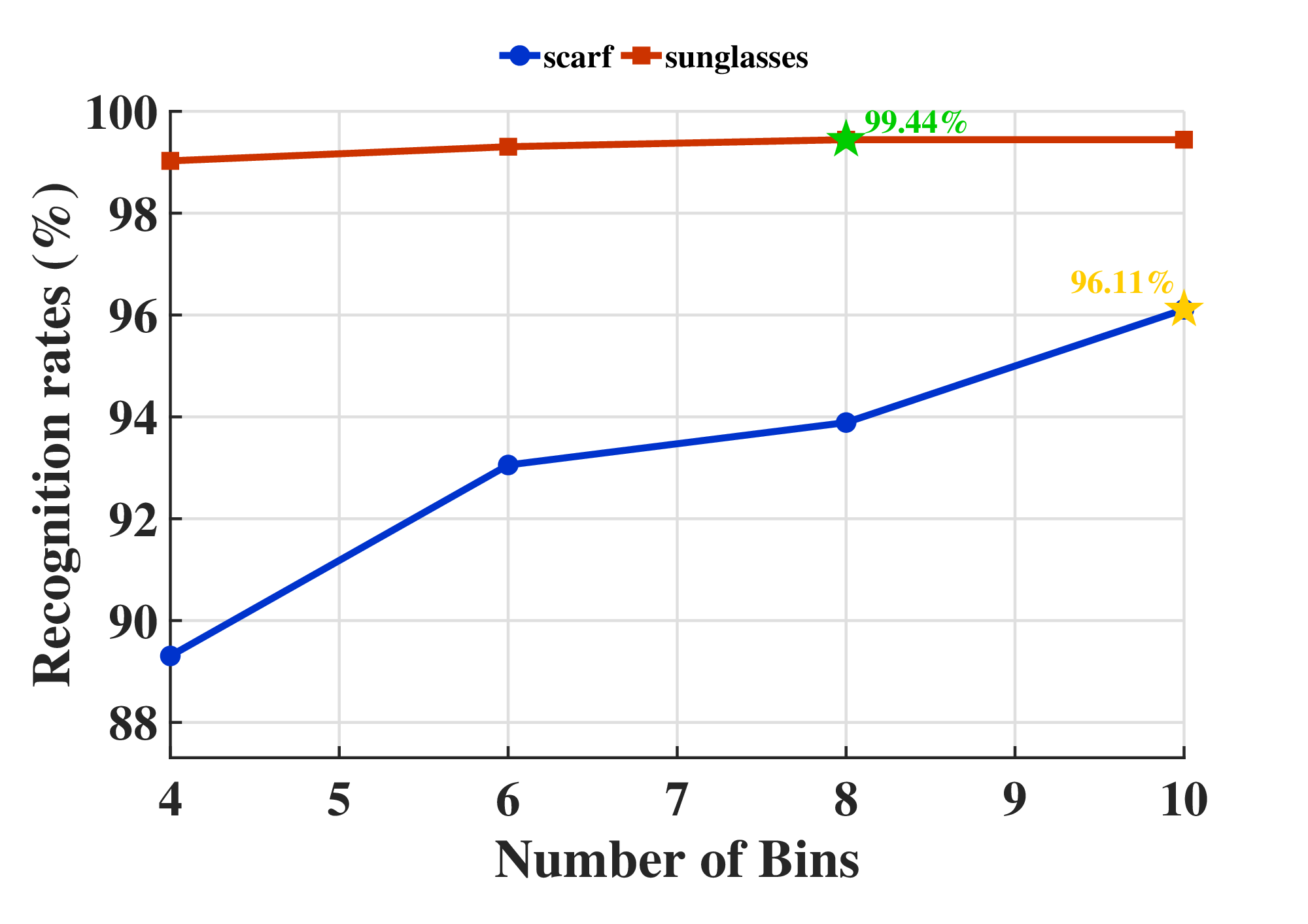}
        \caption{Different number of gradient orientation bins}
    \end{subfigure}
    \hspace{0.01\textwidth}
    \begin{subfigure}{0.48\textwidth}
        \centering
        \includegraphics[width=\linewidth]{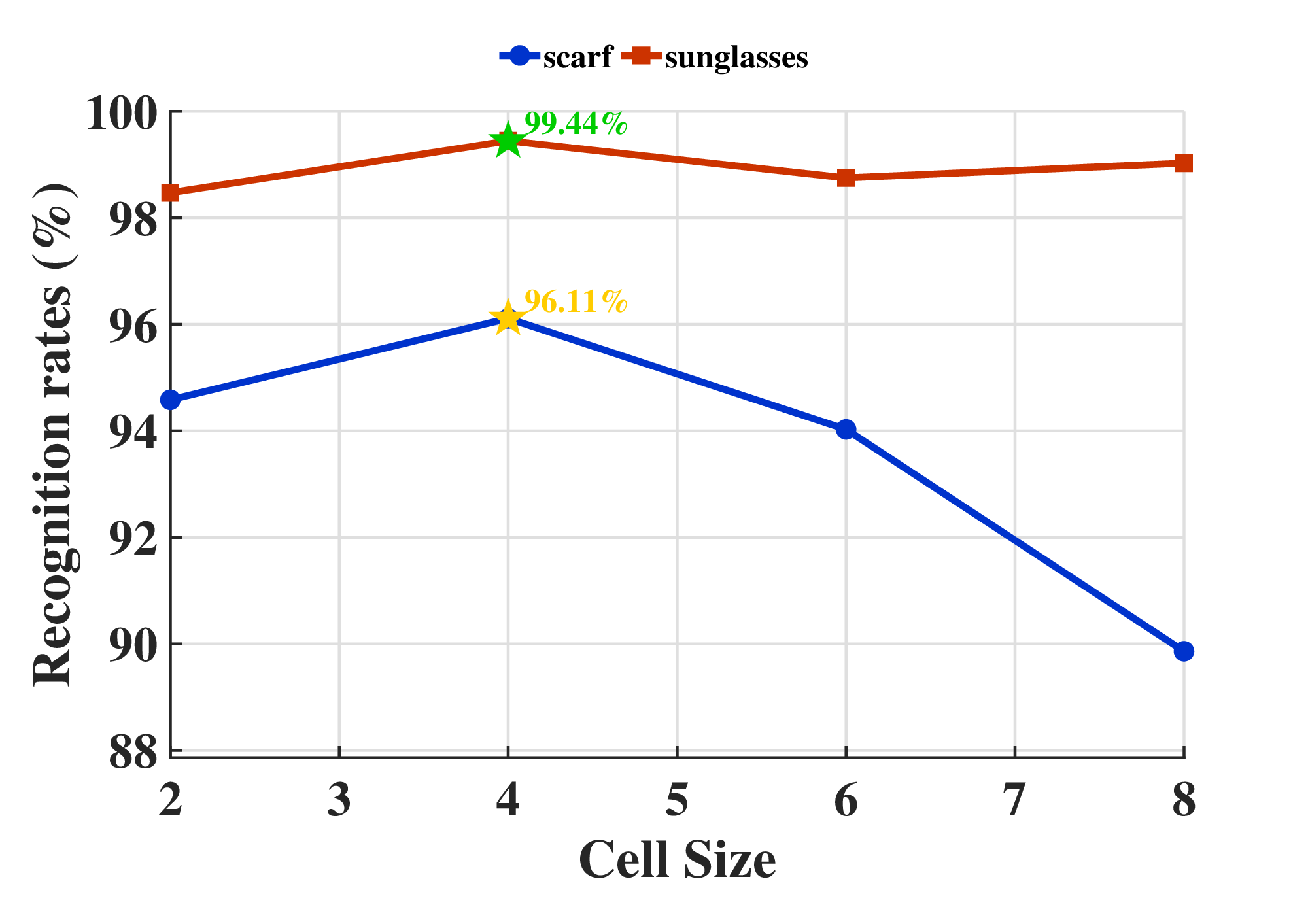}
        \caption{Different cell sizes}
    \end{subfigure}

    \vspace{0.5cm} 

    \begin{subfigure}{0.48\textwidth}
        \centering
        \includegraphics[width=\linewidth]{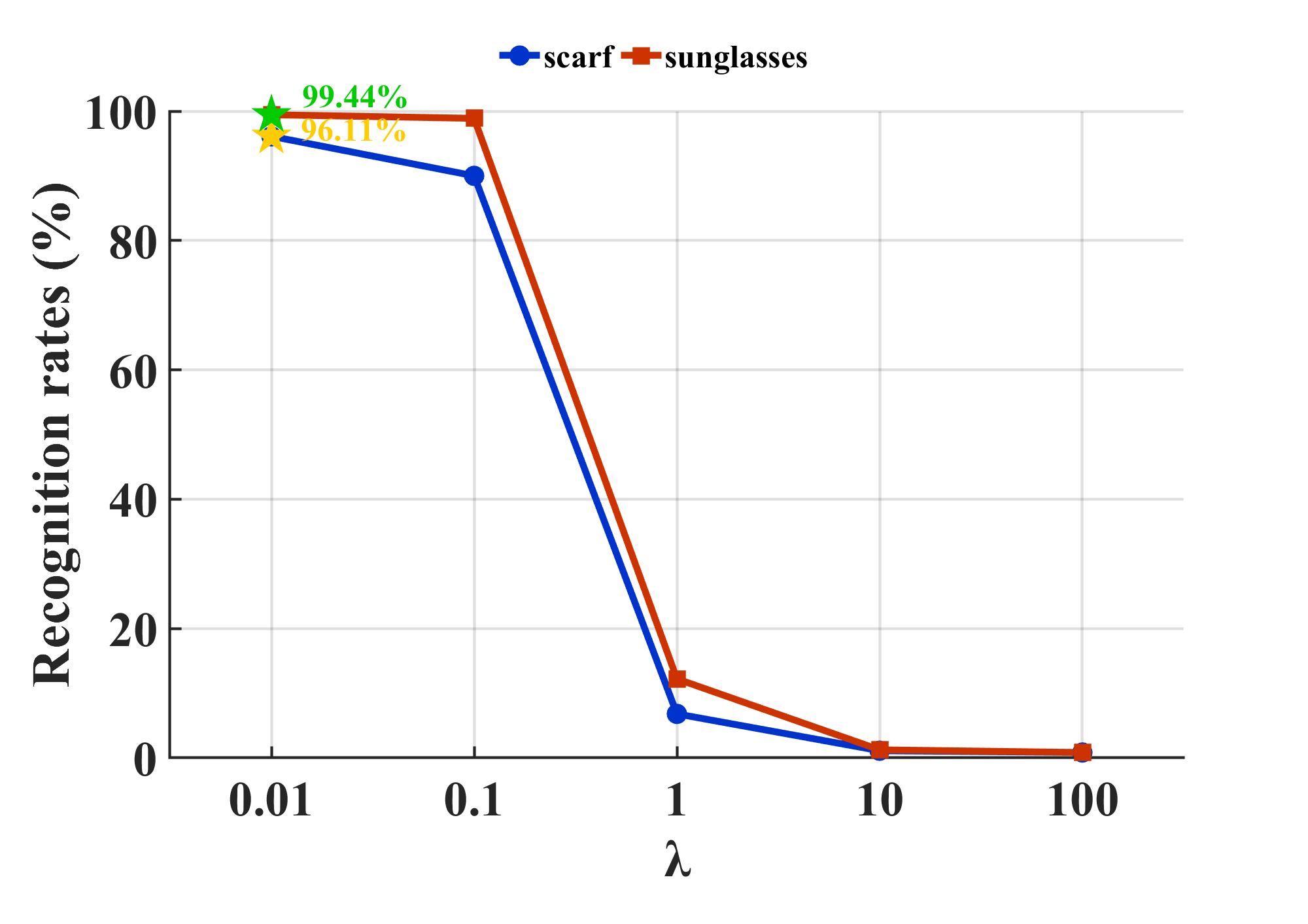}
        \caption{Different regularization parameters $\lambda$}
    \end{subfigure}
    \hspace{0.01\textwidth}
    \begin{subfigure}{0.48\textwidth}
        \centering
        \includegraphics[width=\linewidth]{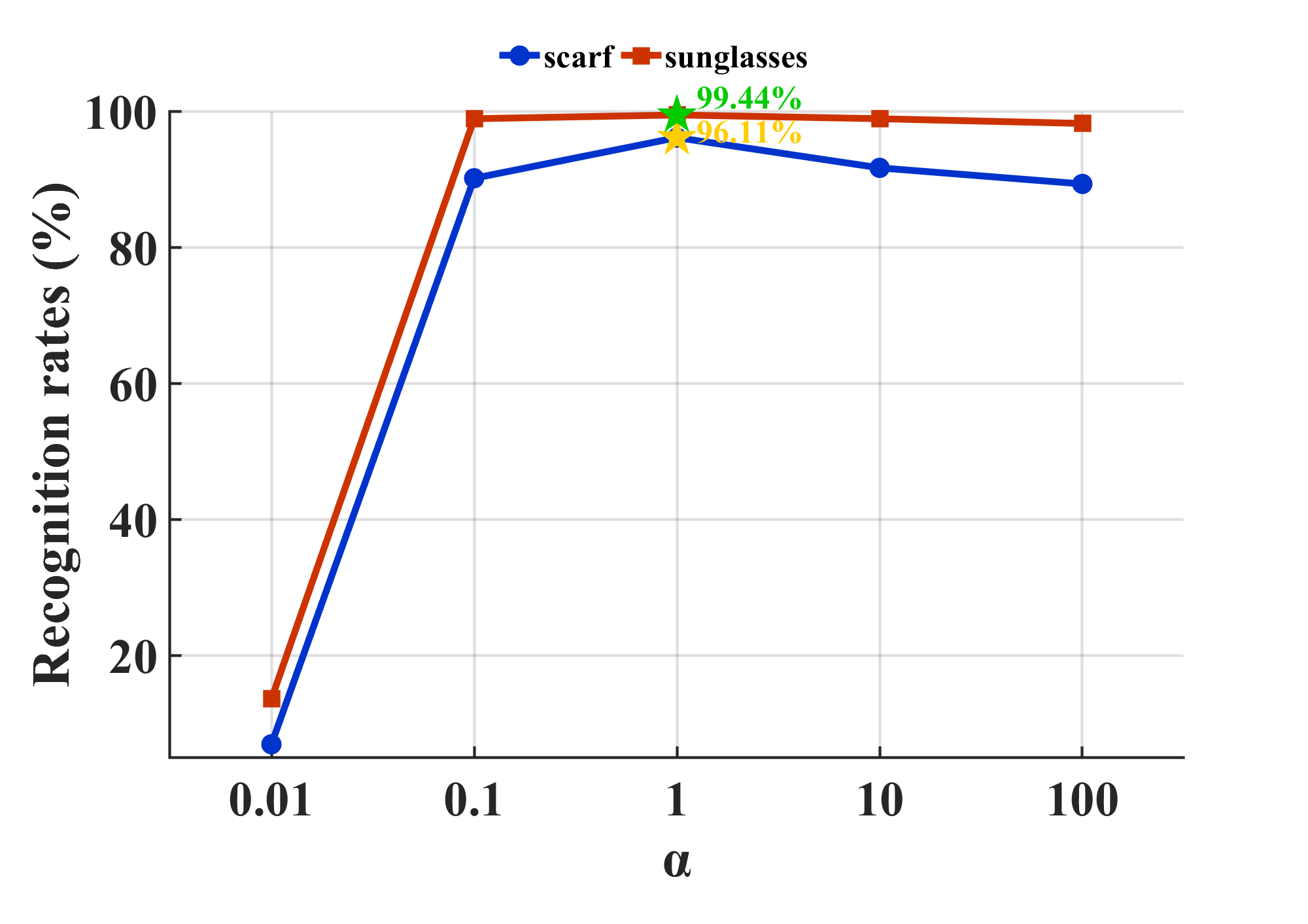}
        \caption{Different regularization parameters $\alpha$}
    \end{subfigure}

    \caption{Recognition rates (\%) with different parameters on the AR dataset}\label{fig:recognition rates of different parameters on AR data}
\end{figure}

\begin{figure}[H]
    \centering
    \begin{subfigure}{0.48\textwidth}
        \centering
        \includegraphics[width=\linewidth]{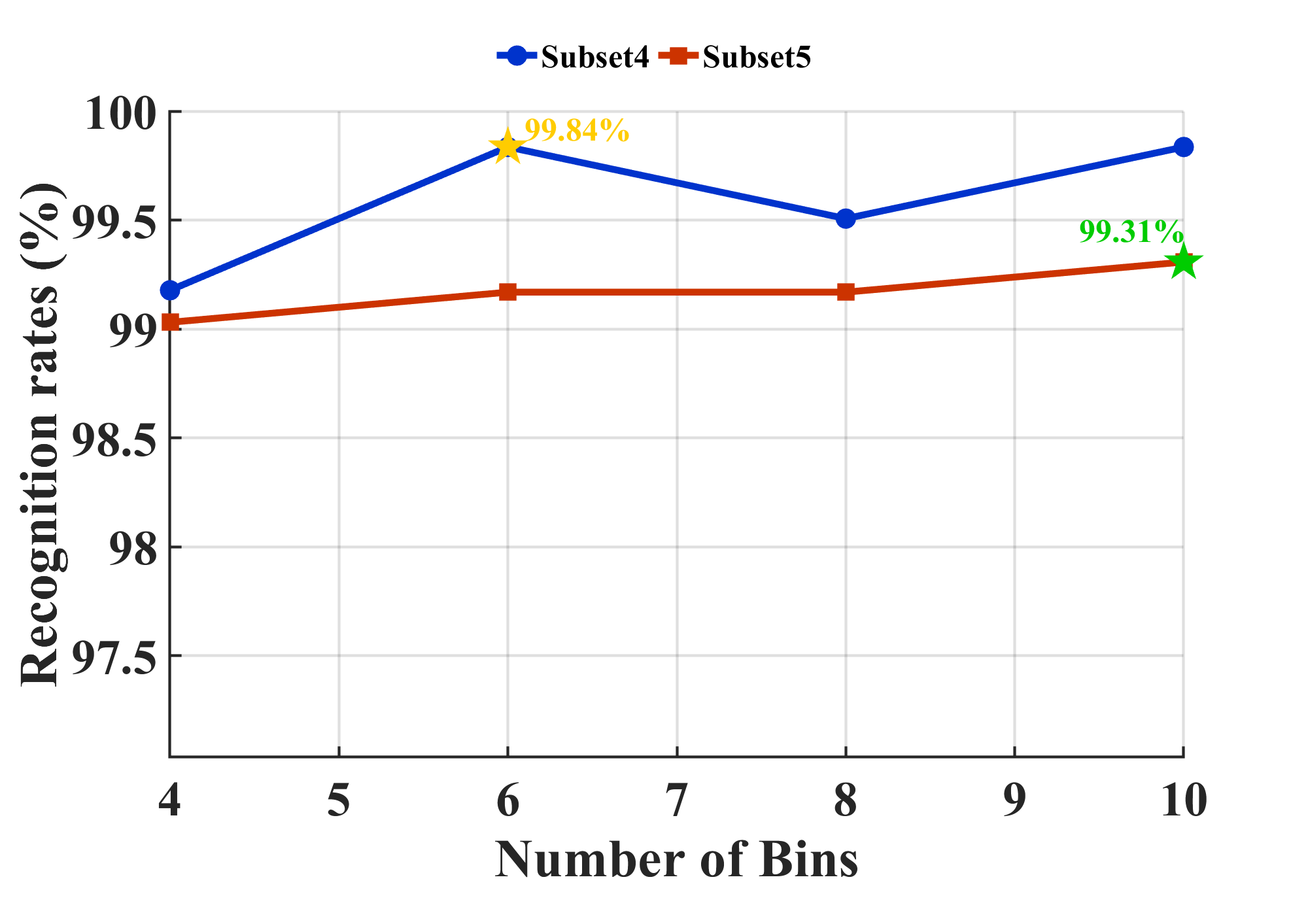}
        \caption{Different number of gradient orientation bins}
    \end{subfigure}
    \hspace{0.01\textwidth}
    \begin{subfigure}{0.48\textwidth}
        \centering
        \includegraphics[width=\linewidth]{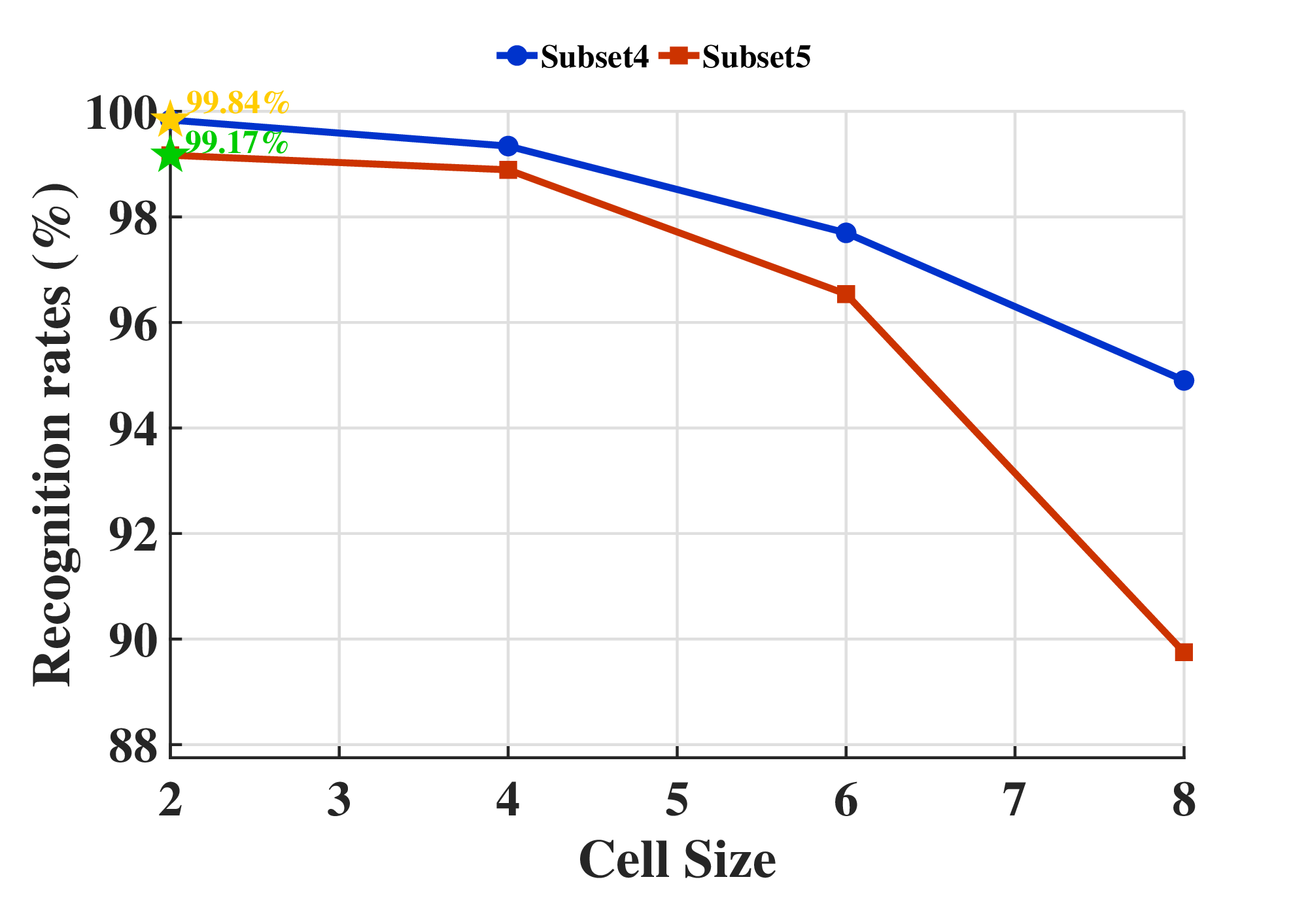}
        \caption{Different cell sizes}
    \end{subfigure}

    \vspace{0.5cm} 

    \begin{subfigure}{0.48\textwidth}
        \centering
        \includegraphics[width=\linewidth]{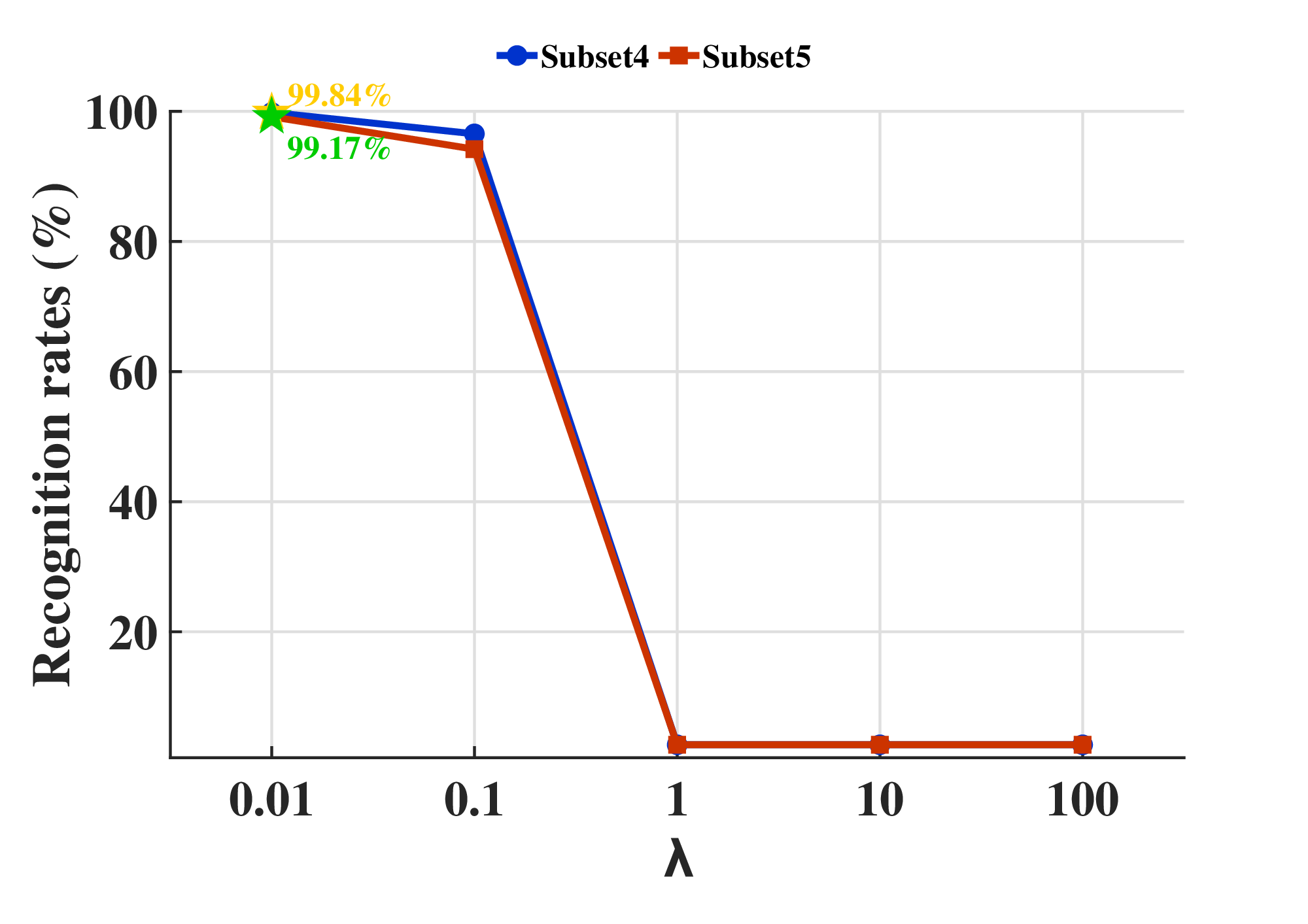}
        \caption{Different regularization parameters $\lambda$}
    \end{subfigure}
    \hspace{0.01\textwidth}
    \begin{subfigure}{0.48\textwidth}
        \centering
        \includegraphics[width=\linewidth]{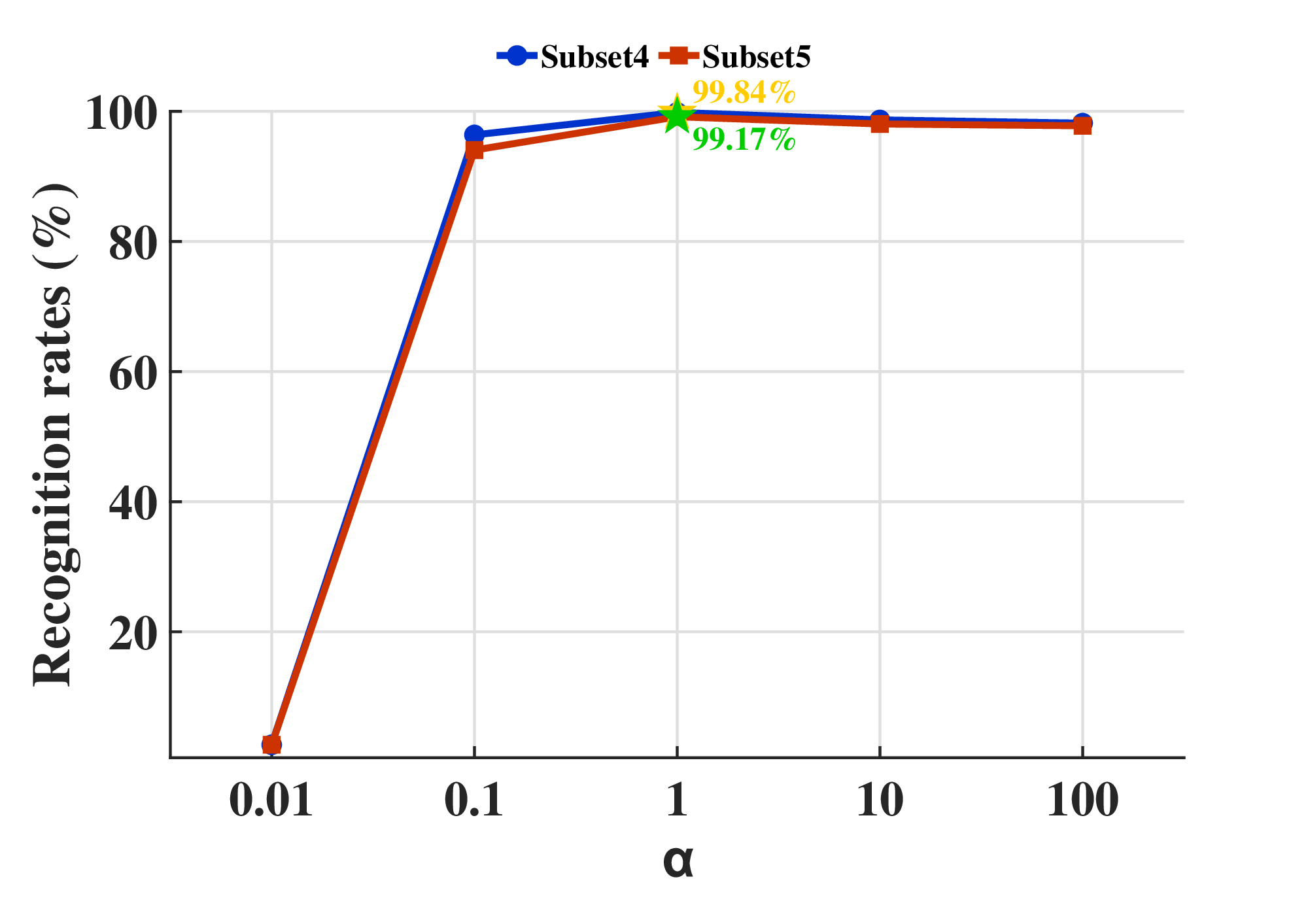}
        \caption{Different regularization parameters $\alpha$}
    \end{subfigure}

    \caption{Recognition rates (\%) with different parameters on the EYB dataset}\label{fig:recognition rates of different parameters on EYB data}
\end{figure}

\begin{figure}[H]
    \centering
    \begin{subfigure}{0.45\textwidth}
        \centering
        \includegraphics[width=\linewidth]{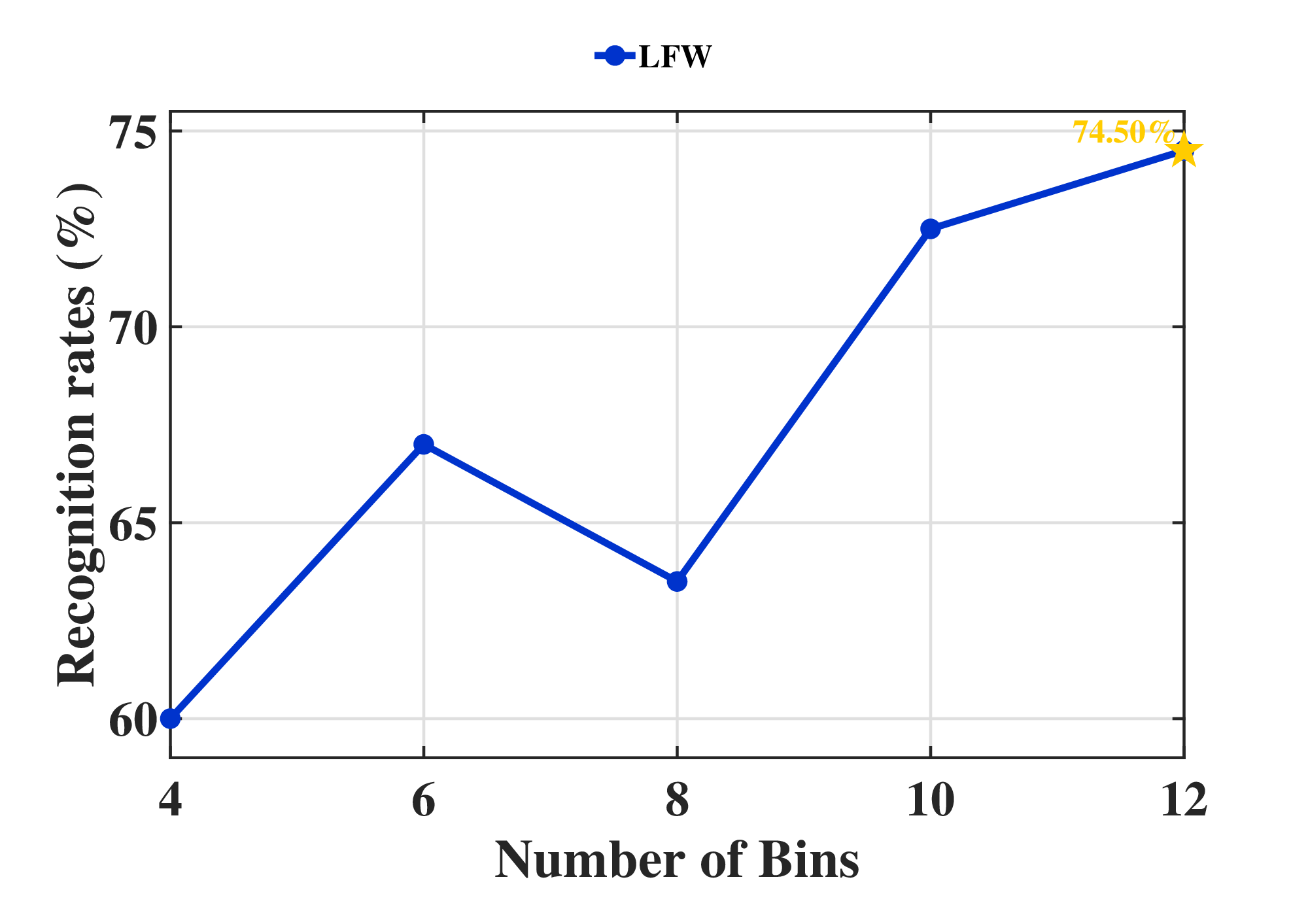}
        \caption{Different number of gradient orientation bins}
    \end{subfigure}
    \hspace{0.01\textwidth}
    \begin{subfigure}{0.45\textwidth}
        \centering
        \includegraphics[width=\linewidth]{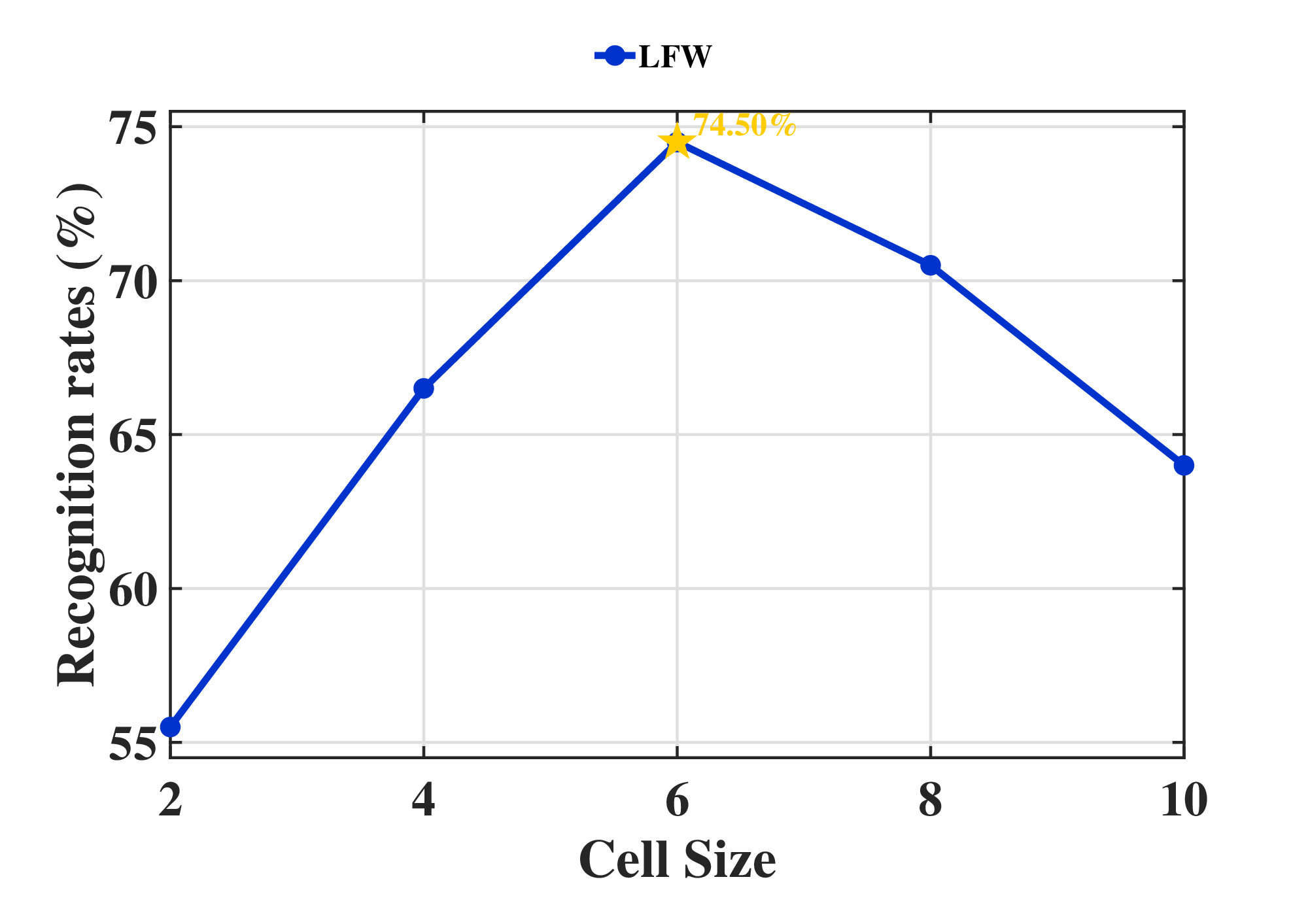}
        \caption{Different cell sizes}
    \end{subfigure}

    \vspace{0.5cm} 

    \begin{subfigure}{0.45\textwidth}
        \centering
        \includegraphics[width=\linewidth]{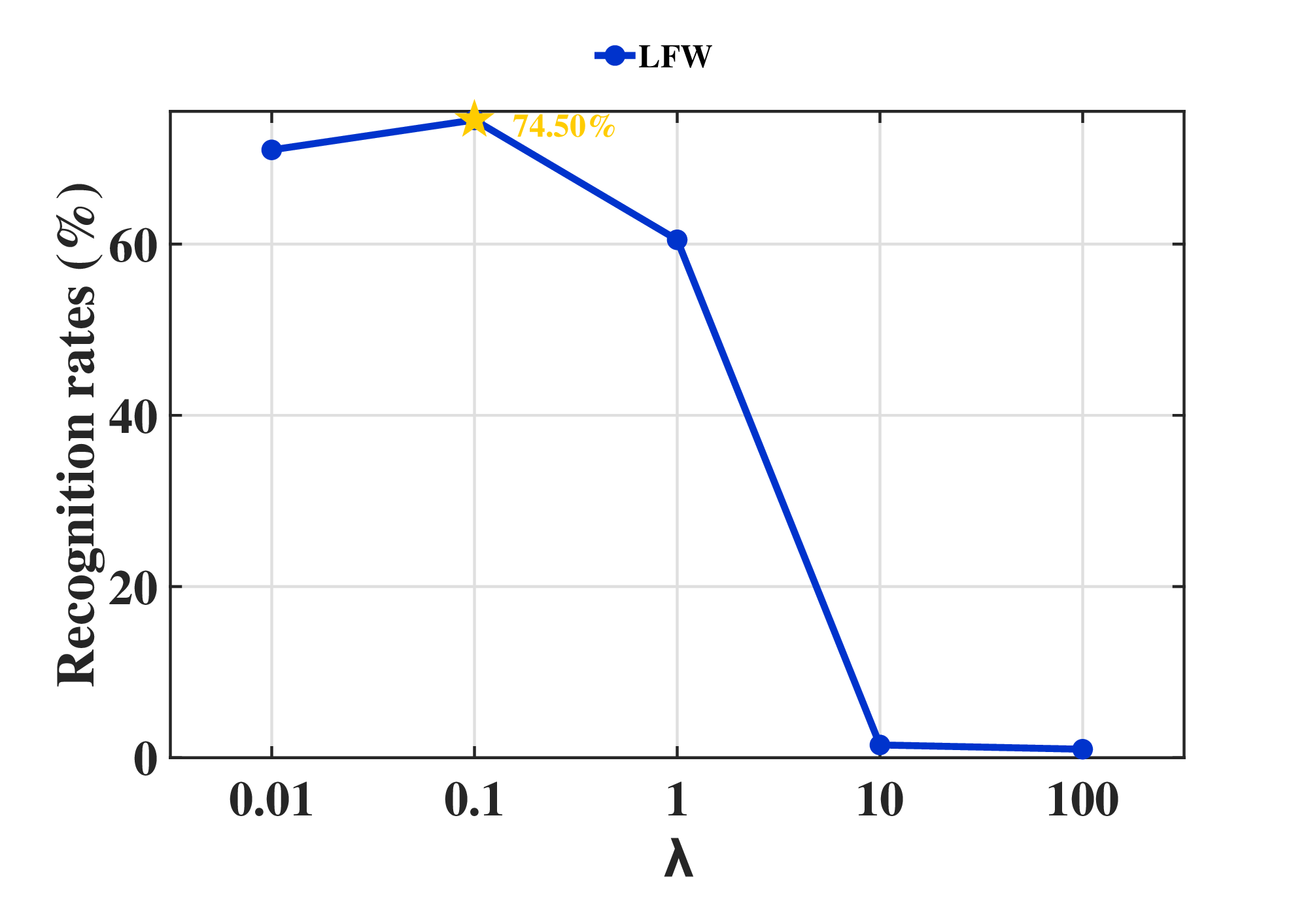}
        \caption{Different regularization parameters $\lambda$}
    \end{subfigure}
    \hspace{0.01\textwidth}
    \begin{subfigure}{0.45\textwidth}
        \centering
        \includegraphics[width=\linewidth]{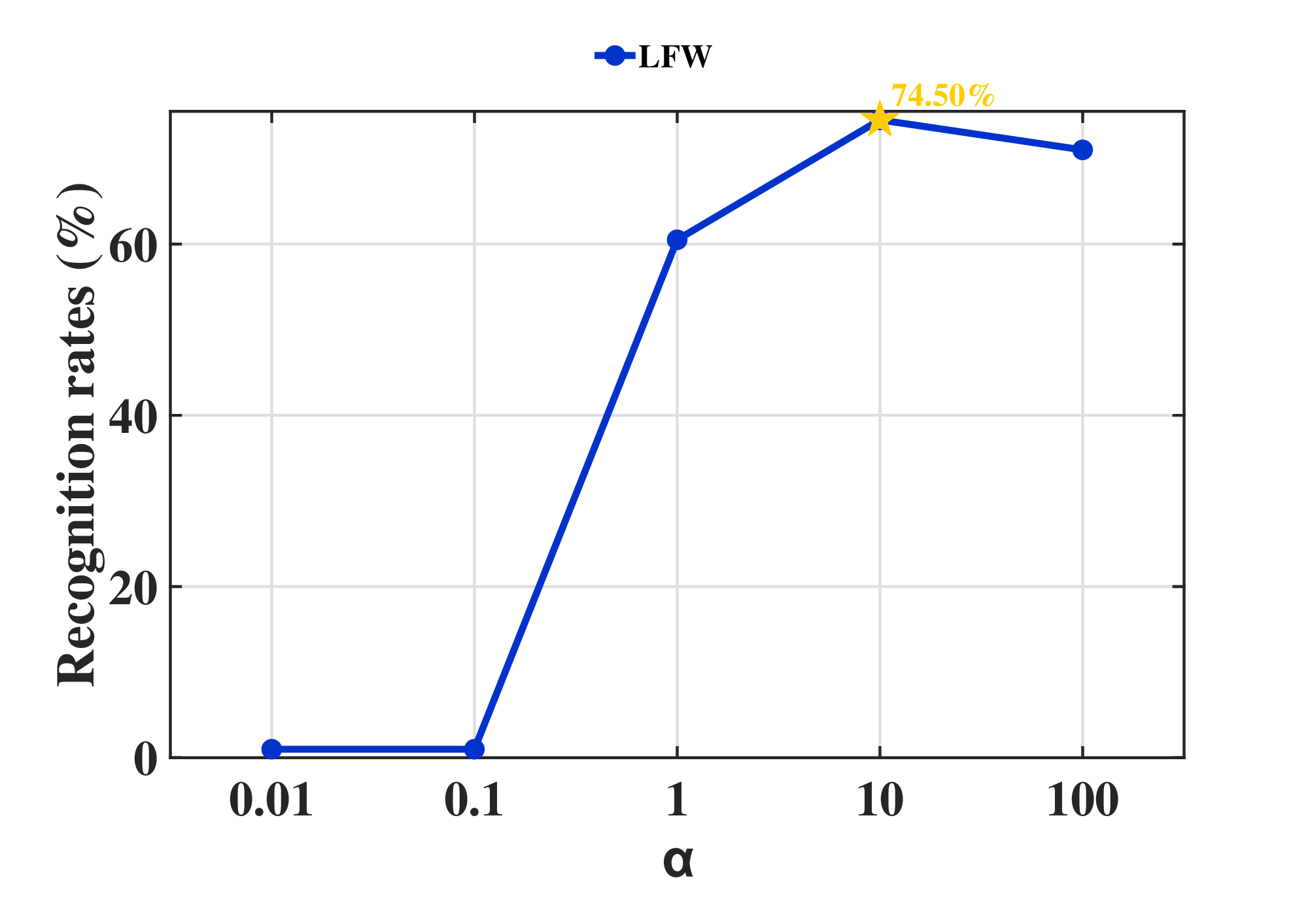}
        \caption{Different regularization parameters $\alpha$}
    \end{subfigure}

    \caption{Recognition rates (\%) with different parameters on the LFW dataset}\label{fig:recognition rates of different parameters on the LFW dataset}
\end{figure}

In this section, we investigate the sensitivity of H2H-GLRSR to its main parameters, including the cell size $N$, the number of orientation bins $B$, and the regularization weights $\lambda$ and $\alpha$. For each dataset, a grid search is performed over a predefined set of candidate values, and Figs.~\ref{fig:recognition rates of different parameters on AR data}-\ref{fig:recognition rates of different parameters on the LFW dataset} illustrate the recognition rates under different parameter settings. As shown in Fig.~\ref{fig:recognition rates of different parameters on AR data}, different occlusion types correspond to different optimal parameter configurations. Sunglasses occlusion achieves a recognition rate of 96.11\% when the number of gradient orientation bins B is 10, while scarf occlusion performs best when B is 8, reflecting the differentiated requirements of different occlusion characteristics for gradient orientation resolution. Meanwhile, the low-rank regularization parameter $\alpha$ being 1 is crucial for improving the recognition rate of scarf occlusion, increasing it from 87\% to 96.11\%. The results in Fig.~\ref{fig:recognition rates of different parameters on EYB data} show that the model exhibits good stability to parameter changes. The parameter $\lambda$ shows a wide optimal range. When $\lambda$ ranges from 0.01 to 0.1, the model maintains excellent performance on both Subset4 and Subset5. Also, when $\alpha$ is 1, the model achieves the best performance under extreme lighting conditions, verifying the effectiveness of the low-rank constraint for illumination modeling. From Fig.~\ref{fig:recognition rates of different parameters on the LFW dataset}, it is not difficult to see that when the number of cells N is 6, the number of gradient orientation bins B is 12, the regularization parameter $\lambda$ is 0.1, and the regularization parameter $\alpha$ is 10, the proposed method can achieve good performance.

\subsection{Computation time}
The above methods are all implemented in Matlab R2024b. The computational experiments were carried out on a 64-bit workstation equipped with an Intel Core i9-13900K processor operating at 3.00 GHz, 96 GB of system memory, and an Intel UHD Graphics 770 integrated GPU. The machine was configured with approximately 11.85 TB of available storage. 

 The experimental results on running time, as summarized in Table~\ref{tab:Running time}, demonstrate the superior computational efficiency of the proposed H2H-GLRSR method. H2H-GLRSR achieves significantly faster processing speeds across all tested datasets (AR, Extended Yale, and LFW) compared to the competing methods SR\_NMR and ID-NMR. Specifically, it runs approximately 80 times faster than ID-NMR and nearly 10 times faster than SR\_NMR on the AR dataset, while maintaining a substantial speed advantage (1.7x to 4.5x faster than competitors) on the Extended Yale subsets. Most notably, on the LFW dataset, H2H-GLRSR completes the task in only 9.01 seconds, which is over 19 times and 30 times faster than SR\_NMR and ID-NMR, respectively. 

Overall, the average computation time of H2H-GLRSR is much lower than that of the comparison methods, which is mainly attributed to: the efficient parallel computation of H2H features, the simplicity of the ADMM optimization algorithm, and the reduction of repeated computation by the global low-rank constraint. The experimental results show that this method can achieve significant improvement in computational efficiency in various scenarios while maintaining excellent recognition performance.

\begin{table}[htbp]
  \centering
  \caption{Running time (second) of SR\_NMR, ID-NMR and H2H-GLRSR methods.}
  \label{tab:Running time}
  \begin{tabular}{l l l l l l}
    \toprule
    \textbf{Methods/Cases} & \multicolumn{2}{c}{\textbf{AR}} & \multicolumn{2}{c}{\textbf{Extend Yale}} & \textbf{LFW} \\
    \cmidrule(lr){2-3} \cmidrule(lr){4-5}
     & sunglasses & scarf & subset4 & subset5 & \\
    \midrule
    SR\_NMR & 244.45 & 242.97 & 295.86 & 343.79 & 175.05 \\
    ID-NMR & 2318.72 & 2334.49 & 500.50 & 796.06 & 276.54 \\
    \textbf{H2H-GLRSR} & \textbf{25.08} & \textbf{28.18} & \textbf{109.53} & \textbf{200.25} & \textbf{9.01} \\
    \bottomrule
  \end{tabular}
\end{table}

 \section{Conclusion}
 We address the challenge of face recognition in complex environments by proposing a hybrid feature descriptor H2H that integrates first-order and second-order gradient information, and on this basis, constructs a Global Low-Rank Sparse Regression model H2H-GLRSR. The H2H feature effectively combines the edge contour information of HOG with the curvature texture information of HOH through parallel computation on a unified grid and early fusion, significantly enhancing the discriminative capability of local image structures. The H2H-GLRSR model introduces a global low-rank constraint on the residual matrix, effectively mining the global correlations of structured noise and occlusion, greatly improving the model's robustness to complex variations. Experimental results on three public datasets, AR, Extended Yale B and LFW, show that the proposed method achieves optimal recognition performance under occlusion, illumination changes, and unconstrained environments. More importantly, the optimization algorithm designed in this paper demonstrates excellent computational efficiency while ensuring leading performance. This favorable balance between performance and efficiency endows the method with considerable potential for practical applications. Future work will focus on further optimizing the feature fusion strategy and algorithm convergence speed to promote the deployment and application of this method in real-time systems.

\section*{Acknowledgments} 
The authors are grateful to the anonymous referees for their valuable comments and suggestions which improved the quality of this paper. This work was supported by National Natural Science Foundation of China (Grant Nos. 12001363 and 72171170), National Key Research and Development Program of China (2023YFC3306100, 2023YFC3306105, 2023YFC3306103), Shanghai Municipal Education and Health Work Committee System United Front Investigation and Research Project (JWTZDY-2025-A15).

\bibliographystyle{elsarticle-num}
\bibliography{references} 

\end{document}